\documentclass[runningheads]{llncs}

\usepackage[mobile]{eccv}

\newif\ifarxiv
\arxivtrue
\usepackage[pagebackref=false,breaklinks=true,colorlinks,citecolor=eccvblue,bookmarks=true,bookmarksnumbered=true]{hyperref}
\hypersetup{
  pdftitle={Event-based Mosaicing Bundle Adjustment},
  pdfsubject={Computer Vision, Robotics, State Estimation},
  pdfauthor={Shuang Guo, Guillermo Gallego},
  pdfkeywords={Event Cameras, Motion Estimation, Bundle Adjustment, Asynchronous Sensor, High Dynamic Range, High Temporal Resolution}
}

\renewcommand{\paragraph}[1]{
 \textbf{#1.} 
}

\usepackage[dvipsnames]{xcolor}
\usepackage{graphicx}
\usepackage{booktabs}

\usepackage{threeparttable} % tablenotes
\usepackage{makecell}
\usepackage{multirow}
\usepackage{rotating}

\usepackage{siunitx}
\usepackage{adjustbox} % resize width if needed

\usepackage{url}
\def\pol{s} % event polarity when the map point is \bp so close to pol\equiv pk
\def\prtl#1#2{\frac{\partial#1}{\partial#2}}

\def\numEvents{N_e} % number of events
\def\numPixels{N_p} % number of pixels
\def\numPoses{N_{\text{poses}}} % number of control poses
\def\Warp{\mathbf{W}}

\def\bx{\mathbf{x}}

\def\Rot{\mathtt{R}}

\def\angvel{\boldsymbol{\omega}} % angular velocity
\def\imgpoint{\bx}

\def\bz{\mathbf{z}}
\def\mJ{\mathtt{J}}
\def\be{\mathbf{e}}%
\def\bb{\mathbf{b}}%
\def\Kint{\mathtt{K}}%
\def\balpha{\boldsymbol{\alpha}}%
\def\bbeta{\boldsymbol{\beta}}%
\def\rotperturb{\boldsymbol{\delta\phi}}%
\def\mE{\mathtt{E}}%
\def\br{\mathbf{r}}%
\def\bc{\mathbf{c}}%
\def\bv{\mathbf{v}}%

\def\Real{\mathbb{R}} % Real numbers

\def\mA{\mathtt{A}}

\def\mZero{\mathtt{0}}
\def\mOne{\mathtt{1}}

\def\bp{\mathbf{p}}
\def\bP{\mathbf{P}}

\def\loss{g} % careful if there is L^2 norm on the same document

\def\NA{\text{N/A}}
\def\dash{\text{-}}

\newcommand{\novalue}{{\textendash}}

\newcommand{\gred}[1]{#1}
\definecolor{light-gray}{gray}{0.6}
\newcommand\gframe[1]{{\color{light-gray}\frame{#1}}}
\newcommand\gframeRed[1]{{\color{red}\frame{#1}}}
\newcommand\gframeGreen[1]{{\color{green}\frame{#1}}}
\newcommand\gframeBlue[1]{{\color{blue}\frame{#1}}}

\def\playroom{\emph{playroom}}
\def\bicycle{\emph{bicycle}}
\def\city{\emph{city}}
\def\street{\emph{street}}
\def\town{\emph{town}}
\def\bay{\emph{bay}}

\def\shapes{\emph{shapes}}
\def\poster{\emph{poster}}
\def\boxes{\emph{boxes}}
\def\dynamic{\emph{dynamic}}

\def\bridge{\emph{bridge}}
\def\crossroad{\emph{crossroad}}
\def\graffiti{\emph{graffiti}}

\def\atrium{\emph{atrium}}

\def\esmt{EKF-SMT}
\def\psmt{PF-SMT}
\def\rtpt{RTPT}
\def\cmaxgae{CMax-GAE}
\def\cmaxw{CMax-$\angvel$}

\def\LM{Levenberg-Marquardt}

\usepackage{pifont}% http://ctan.org/pkg/pifont
\newcommand{\yesmark}{\ding{51}}% good mark
\newcommand{\nomark}{\ding{55}}% bad mark

\def\emptycol{\hspace{1ex}} % spacing empty cols

\def\Ltwo{L^2}

\usepackage{orcidlink}

\ifarxiv
\usepackage[absolute]{textpos}
\fi

\begin{document}

% ---------------------------------------------------------------
\title{Event-based Mosaicing Bundle Adjustment} 

\ifarxiv
\definecolor{somegray}{gray}{0.5}
\newcommand{\darkgrayed}[1]{\textcolor{somegray}{#1}}
\begin{textblock}{11}(2.5, -0.1)  % {hsize}(hpos,vpos)
\begin{center}
\darkgrayed{This paper has been accepted for publication at the European Conference on Computer Vision (ECCV), 2024. \copyright Springer}
\end{center}
\end{textblock}
\fi 

\titlerunning{Event-based Mosaicing Bundle Adjustment}

\author{Shuang Guo\inst{1}\orcidlink{0000-0002-0142-0678} \and
Guillermo Gallego\inst{1,2}\orcidlink{0000-0002-2672-9241}}

\authorrunning{S.~Guo and G. Gallego}

\institute{TU Berlin and Robotics Institute Germany, Berlin, Germany \and
Einstein Center Digital Future and SCIoI Excellence Cluster, Berlin, Germany
% \email{shuang.guo@campus.tu-berlin.de} \quad
% \email{guillermo.gallego@tu-berlin.de}
}

\maketitle

\begin{abstract}
We tackle the problem of mosaicing bundle adjustment (i.e., simultaneous refinement of camera orientations and scene map) for a purely rotating event camera.
We formulate the problem as a regularized non-linear least squares optimization.
The objective function is defined using the linearized event generation model in the camera orientations and the panoramic gradient map of the scene.
We show that this BA optimization has an exploitable block-diagonal sparsity structure, so that the problem can be solved efficiently.
To the best of our knowledge, this is the first work to leverage such sparsity to speed up the optimization in the context of event-based cameras, 
without the need to convert events into image-like representations.
We evaluate our method, called EMBA, on both synthetic and real-world datasets to show its effectiveness (50\% photometric error decrease), yielding results of unprecedented quality.
In addition, we demonstrate EMBA using high spatial resolution event cameras, yielding delicate panoramas in the wild, even without an initial map.
Project page: \url{https://github.com/tub-rip/emba}.
\end{abstract}
\section{Introduction}
\label{sec:intro}

% 1. What is the problem?
% 2. Why is it important?
% 3. Why is the problem hard? What makes it challenging?
% 4. How far has existing work come? What is the frontier?
% 5. Why hasn't the problem been solved? What is the stumbling block?
% 6. What does our paper contribute?
% 7. What is the key idea? What is the magic trick? What is the new insight or technique that enables us to advance the frontier?
% 8. What do the experiments say?

% There are five basic questions we ask to authors (IJRR):
% "Why is this problem important?", 
% "Why wasn't it solved before?", 
% "What's the key idea in the solution?", 
% "How do you show that it really works?", and 
% "How can others use your results?".
% Results will have to be demonstrated by all relevant and applicable scientific means - be it mathematical proofs, statistically significant and reproducible experimental tests, demos and field demonstrations, or whatever may be needed to convince a duly skeptical, critical scientist.
% ------------------------------------------------------------------------------------------

% Introduce event cameras
Event cameras are novel bio-inspired visual sensors that measure per-pixel brightness changes \cite{Lichtsteiner08ssc,Son17isscc,Finateu20isscc}. 
In contrast to the images/frames captured by standard cameras, the output of an event camera is a stream of asynchronous events.
This unique working principle endows event cameras with great potential in the tasks of camera motion estimation and scene reconstruction, especially in scenarios of high dynamic range (HDR), low power consumption and/or fast motion \cite{Gallego20pami}.

% Introduce the problem of bundle adjustment
Bundle Adjustment (BA) is the problem of jointly refining the camera motion and the reconstructed scene map 
that best fit the visual data through a given objective function \cite{Triggs00,Alismail16accv} (e.g., reprojection or photometric error).
It is a paramount topic in photogrammetry, computer vision and robotics,
enabling accurate positioning and measurement technology for applications such as image stitching \cite{Brown06ijcv}, visual odometry (VO) \cite{Engel17pami}, simultaneous localization and mapping (SLAM) \cite{Alismail16accv,Cadena16tro} and AR/VR \cite{Engel23aria}.
BA with frame-based cameras is a mature topic \cite{Hartley03book,Szeliski10book,Alismail16accv,Klenk19thesis}.
% Point out the gap in current works
In contrast, BA with event cameras is still in its infancy, which limits the maturity of the above-mentioned applications for event cameras.

% Why this problem is hard?
A key problem for BA and SLAM-related tasks is data association, i.e., establishing correspondences between measurements to identify which pixels observe the same scene point \cite{Gehrig19ijcv,Hidalgo22cvpr}.
So far, event-based BA has been applied in a feature-based (i.e., indirect) manner, i.e., extracting sparse keypoints from image-like event representations and associating them over time (e.g., \cite{Chin19cvprw,Wang23ral}).
However, this discards the large amount of information contained in the events (as shown in image reconstruction \cite{Rebecq19pami,Mostafavi21ijcv,Zhang22pami,Stoffregen20eccv}) and/or quantizes their high temporal resolution.
Instead, recent development in direct methods with event cameras \cite{Reinbacher17iccp,Hidalgo22cvpr,Guo24tro} suggest that it should be possible to achieve BA while exploiting the unique characteristics of events, 
namely that they are continuously (asynchronously) triggered by edges as the camera moves, 
and that each event is a relative brightness measurement (i.e., an increment if using logarithmic scale).

\begin{figure}[t]
    \centering
    \includegraphics[width=\linewidth]{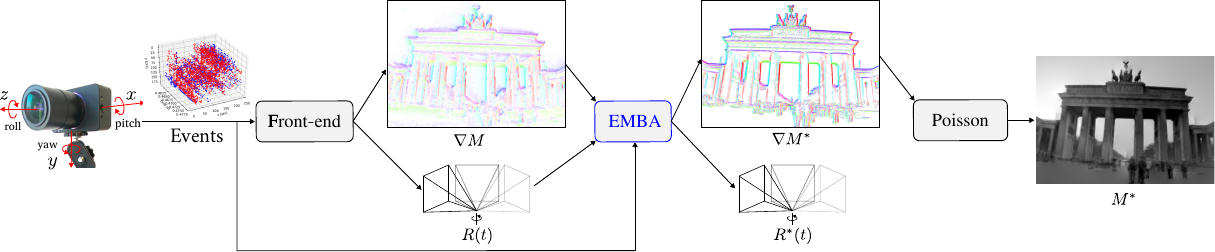}
    \caption{Our back-end module EMBA jointly refines the camera rotations and panoramic gradient map. 
    The intensity map can be recovered by solving Poisson's equation.}
\label{fig:eyecatcher}
\end{figure}

% What do we do in this paper?
This paper proposes an event-based mosaicing bundle adjustment (EMBA) method to tackle the photometric BA problem for event cameras (\cref{fig:eyecatcher}).
Rotational motion is a rich and practical scenario, as shown by previous works \cite{Cook11ijcnn,Kim14bmvc,Gallego17ral,Reinbacher17iccp,Kim21ral,Chin19cvprw,Chin20wacv,Guo24tro}.
It is essential to many applications: panorama creation (e.g., in smartphones), star tracking \cite{Chin20wacv}, and VO/SLAM in dominantly-rotational motion cases (e.g., satellites \cite{Chin19cvprw}).
We leverage the linearized event generation model (LEGM) to formulate the problem as a regularized non-linear least squares (NLLS) optimization in the high-dimensional space of camera motions and panoramic gradient maps.
Due to the sparse property of event data, only a portion of pixels of the panoramic map are observed are consequently refined, which naturally leads to a semi-dense gradient map.
Moreover, the LEGM also yields a block-diagonal sparsity pattern within the system equations, which we leverage to design an efficient second-order solver.

% The outcomes and findings
Therefore, to the best of our knowledge, EMBA is novel.
In the experiments, we run EMBA to refine the camera motion trajectories and maps obtained by four state-of-the-art event-based rotation estimation front-end methods \cite{Kim14bmvc,Reinbacher17iccp,Gallego17ral,Kim21ral}, on both synthetic and real-world datasets.
The results show notable improvements in terms of both camera motion and map quality, revealing previously hidden scene details.
We also demonstrate the application of EMBA to generate high-quality panoramas in outdoor scenes with high resolution event cameras, without requiring an initial map.
That is, EMBA just needs a set of initial camera rotations (e.g., provided by an IMU or some front-end) to recover a delicate panorama from scratch while jointly refining the camera motion.

Our contributions can be summarized as follows:
\begin{itemize}
    \item We propose a novel event-only mosaicing bundle adjustment method, 
    which refines an event-camera's trajectory orientation and gradient map,
    producing a high quality grayscale panoramic map of the scene (\cref{sec:method}).
    Its key ingredients are: formulating the BA problem as a regularized NLLS optimization and leveraging the block-diagonal sparsity pattern induced by the chosen parameterization to implement an efficient solver (\cref{sec:method:sparsity}).
    \item We conduct a comprehensive evaluation on synthetic and real-world datasets (\cref{sec:experim}) using four state-of-the-art front-ends for initialization. 
    We demonstrate the method using high-resolution event cameras (VGA and HD), obtaining remarkable panoramas without map initialization (\cref{sec:experim:wild}).
    \item We make the source code publicly available.
\end{itemize}

\section{Related Work}
\label{sec:related}

\begin{table}[tb]
\centering
\caption{\emph{VO/SLAM systems that use event data}. 
The columns indicate the number of degrees of freedom (DOFs), 
the type of method (\textbf{D}irect or \textbf{I}ndirect --feature-based),
whether there is a refinement step (back-end \cite{Cadena16tro}),
and whether the method exploits the event generation model (linearized --LEGM-- or not).
\label{tab:eslam:methods}
}
\begin{adjustbox}{max width=1.0\linewidth}
\setlength{\tabcolsep}{3pt}
\begin{tabular}{lcccccl}
\toprule
\textbf{System} & \textbf{Year} & \textbf{DOFs} & \textbf{Refine} & \textbf{D/I} & \textbf{EGM} & \textbf{Remarks}\\
\midrule
Weikersdorfer et al. \cite{Weikersdorfer13icvs} & 2013 &  3 & \nomark & D & \nomark & Edge map; 2D scenario\\
\psmt{} \cite{Kim14bmvc} & 2014 &  3 & \nomark & D & \yesmark & LEGM. Brightness map\\ 
\rtpt{} \cite{Reinbacher17iccp} & 2017 &  3 & \nomark & D & \nomark & Probabilistic map\\
\cmaxw{} \cite{Gallego17ral} & 2017 &  3 & \nomark & D & \nomark & The map is a local IWE\\
\esmt{} \cite{Kim18phd} & 2018 &  3 & \nomark & D & \yesmark & LEGM. Brightness map\\ 
Chin et al.~\cite{Chin19cvprw} & 2019 &  3 & \yesmark & I & \nomark & Converts events into frames\\
\cmaxgae{} \cite{Kim21ral} & 2021 &  3 & \nomark & D & \nomark & The map is a growing 3D-point set\\
CMax-SLAM \cite{Guo24tro} & 2024 &  3 & \yesmark & D & \nomark & The map is a panoramic IWE\\
\textbf{This work} & 2024 &  3 & \yesmark & D & \yesmark & LEGM. Event-only photometric BA\\
\bottomrule
\end{tabular}
\end{adjustbox}
\end{table}

\Cref{tab:eslam:methods} summarizes some of the VO/SLAM methods operating on event data.

\textbf{Event-based Rotation Estimation}. 
Several works have demonstrated the capabilities of event cameras to estimate rotational motion in challenging scenarios (e.g., high speed, HDR).
Kim \emph{et al} \cite{Kim14bmvc} proposed a 3-DOF simultaneous mosaicing and tracking (SMT) method consisting of two Bayesian filters operating in parallel (\psmt{} -- particle filter SMT); it estimated the camera motion and a grayscale intensity map of the scene.
Later, the tracker was replaced by a Kalman filter \cite{Gallego15arxiv}, yielding \esmt{} \cite{Kim18phd}.
Although EMBA uses a similar measurement model (LEGM) as SMT, the latter only performs local-time estimation since it is filter-based.
Conversely, EMBA is optimization-based, so it can perform global refinement in both time and map domains.
Also working in parallel, a real-time panoramic tracking and probabilistic mapping was presented in \cite{Reinbacher17iccp} (\rtpt{}),
where the panoramic map of the scene stored the spatial event rate at each point (instead of intensity).
Using contrast maximization (CMax), \cite{Gallego17ral} proposed to estimate the camera's angular velocity 
by warping events on the image plane and aligning them via a focus function \cite{Gallego19cvpr}.
The work has been extended to jointly estimate angular velocity and orientation in \cite{Kim21ral} (\cmaxgae{}).

\textbf{Bundle Adjustment}.
All above-mentioned methods are short-term, i.e., front-ends of SLAM systems.
They lack a BA refinement module, i.e., a SLAM back-end \cite{Cadena16tro}, which is desirable to improve accuracy and consistency.
Surveying the literature, \cite{Chin19cvprw} introduced a BA approach for an event-based system; 
but it was feature-based and tested only on synthetic star-tracking data. 
Guo et al. \cite{Guo24tro} augmented \cite{Gallego17ral} with a back-end, but the map was obtained as a by-product of camera trajectory refinement, resulting in a panoramic edgemap (no intensity map).
Expanding the survey to 6-DOF motions, USLAM \cite{Rosinol18ral} fused events, frames and IMU data: keypoints were extracted from motion-compensated event-images and frames, and fed to a classical back-end \cite{Leutenegger15ijrr}.
Recent work \cite{Hidalgo22cvpr} (EDS) proposed an event-aided direct VO system, in which event data was leveraged to track the camera motion during the blind time between frames. 
The system borrowed the photometric BA module from direct methods like DSO \cite{Engel17pami}, which works on images.
Current stereo methods are semi-dense and lack a back-end \cite{Zhou20tro}, or have a back-end but are feature-based (i.e., indirect) \cite{Wang23ral}. 
Therefore, to the best of our knowledge, event-only photometric (i.e., direct) BA is still an unexplored topic, which we address.
\section{Event-based Mosaicing Bundle Adjustment}
\label{sec:method}

\def\figWidth{0.2\linewidth}
\begin{figure*}[t]
	\centering
    {\footnotesize
    \setlength{\tabcolsep}{1pt}
	\begin{tabular}{
    >{\centering\arraybackslash}m{0.3cm}
	>{\centering\arraybackslash}m{0.35\linewidth} 
	  >{\centering\arraybackslash}m{\figWidth}
    >{\centering\arraybackslash}m{\figWidth}
    >{\centering\arraybackslash}m{\figWidth}
    }
    \rotatebox{90}{\makecell{{\scriptsize Initial map}}}
	& \includegraphics[width=\linewidth]{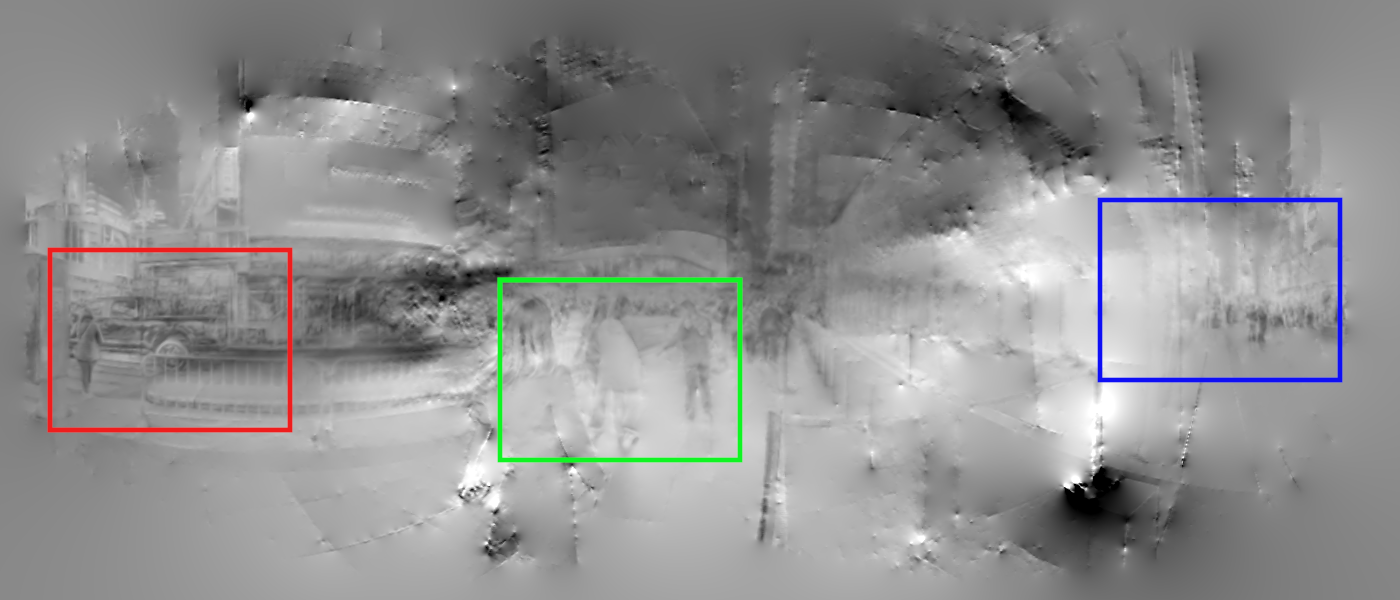}
    & \gframeRed{\includegraphics[width=\linewidth]{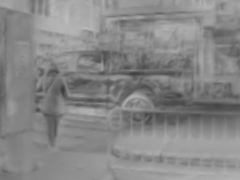}}
    & \gframeGreen{\includegraphics[width=\linewidth]{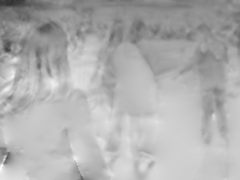}}
    & \gframeBlue{\includegraphics[width=\linewidth]{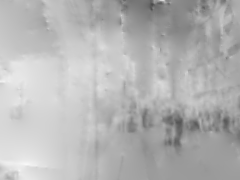}}
    \\[-0.2ex]

    \rotatebox{90}{\makecell{{\scriptsize Refined map}}}
    & \includegraphics[width=\linewidth]{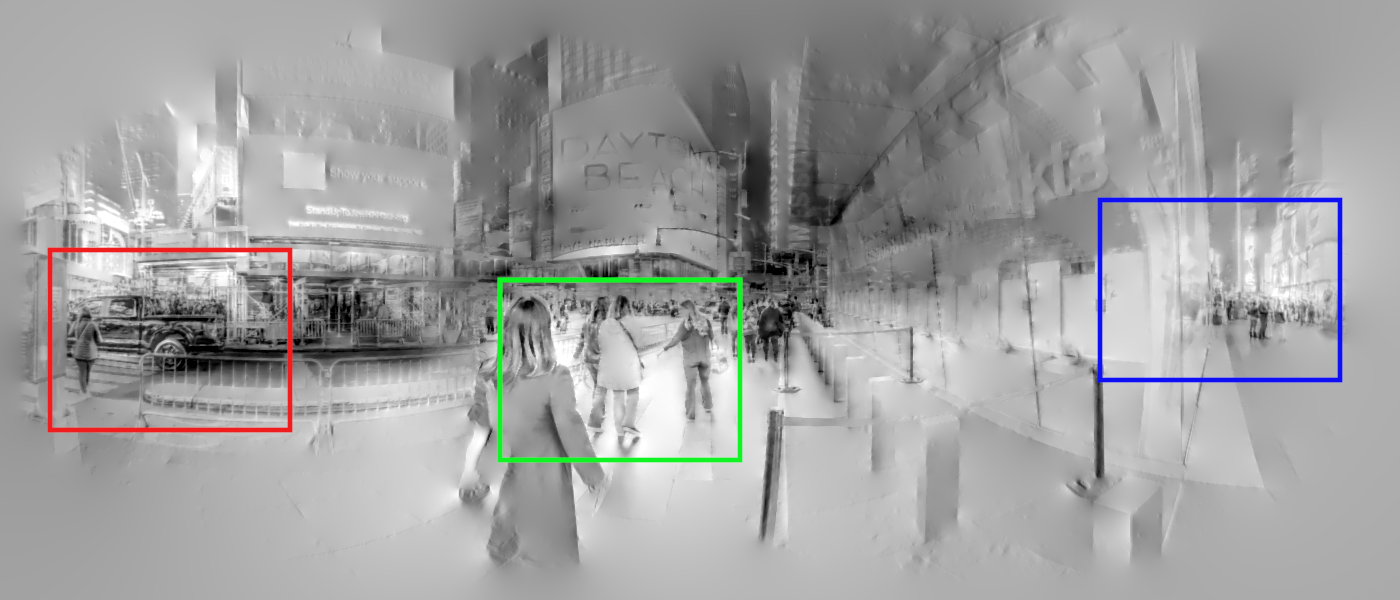}
    & \gframeRed{\includegraphics[width=\linewidth]{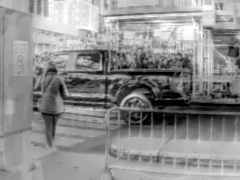}}
    & \gframeGreen{\includegraphics[width=\linewidth]{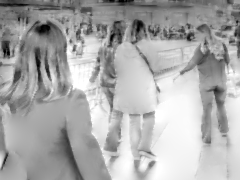}}
    & \gframeBlue{\includegraphics[width=\linewidth]{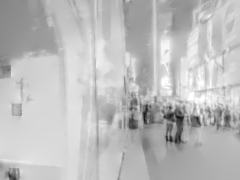}}
    \\[-0.2ex]

    \rotatebox{90}{\makecell{{\scriptsize Refined grad.}}}
    & \gframe{\includegraphics[width=\linewidth]{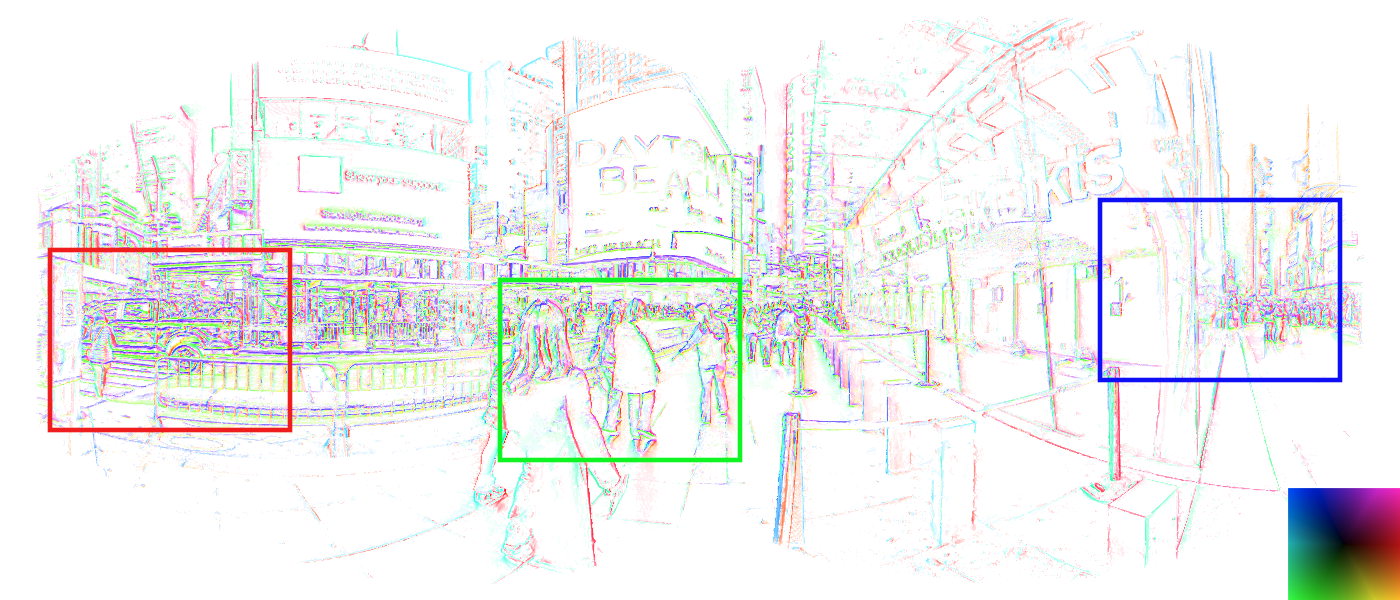}}
    & \gframeRed{\includegraphics[width=\linewidth]{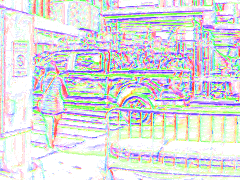}}
    & \gframeGreen{\includegraphics[width=\linewidth]{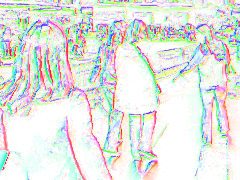}}
    & \gframeBlue{\includegraphics[width=\linewidth]{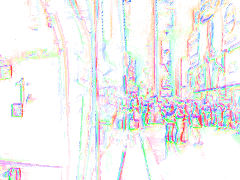}}
    \\
	\end{tabular}
	}
    \caption{Initial intensity map (top), 
    and final map $M$ (middle), 
    via the refined gradient map $\nabla M$ (bottom), 
    for the \street{} data from \cite{Guo24tro}. 
    Three insets are also shown.
    \label{fig:map_comp}
    }
\end{figure*}

\subsection{Event Generation Model (EGM)}
\label{sec:method:EGM}
\textbf{EGM on the sensor}. 
Each pixel of an event camera measures brightness changes independently, 
producing an event $e_k \doteq (\imgpoint_k,t_k,\pol_k)$ when the logarithmic intensity change $\Delta L$ at the pixel reaches a preset contrast threshold $C$ \cite{Gallego20pami}:
\begin{equation}
    \Delta L \doteq L(\imgpoint_k,t_k) - L(\imgpoint_k,t_k - \Delta t_k) = \pol_k C,
    \label{eq:EGM}
\end{equation}
where the event polarity $\pol_k \in \left\{ +1,-1 \right\}$ indicates the sign of the intensity change, and $\Delta t_k$ is the time elapsed since the last event at the same pixel $\imgpoint_k = (x_k,y_k)^\top$.

Assuming brightness constancy (i.e., optical flow constraint), one can further linearize \eqref{eq:EGM} to obtain the ``linearized event generation model'' (\textbf{LEGM}) \cite{Gallego20pami}:
\begin{equation}
    \Delta L \approx - \nabla L(\imgpoint_k,t_k) \cdot \bv \Delta t_k = \pol_k C.
    \label{eq:LEGM}
\end{equation}
It states that the brightness change $\Delta L$ is caused by an edge $\nabla L$ moving with velocity $\bv$ during $\Delta t$ over a displacement $\Delta \imgpoint = \bv \Delta t$.
The dot product captures the condition that no event is triggered if the motion is parallel to the edge.

\textbf{EGM on the scene map}. 
Following \cite{Kim14bmvc}, we may model the scene map using a mosaic, $M: \Real^2 \to \Real$ (e.g., \cref{fig:map_comp}), 
where each map point $\bp$ holds the logarithmic intensity of the 3D world point viewed in the direction of $\bp$.
As the camera rotates, the correspondence between camera pixels $\imgpoint$ and map points $\bp$ varies. 
This warp (i.e., geometric transformation) depends on the camera orientation $\Rot(t)$, intrinsic calibration $\Kint$ and type of projection model $\pi$ (e.g., equirectangular) used to represent the map: $\bx\mapsto\bp$, i.e., 
\begin{equation}
    \bp(t) \doteq \Warp(\bx;\Rot(t),\Kint,\pi).
\end{equation}
Given this correspondence, we may reformulate \eqref{eq:LEGM} in terms of the map:
\begin{equation}
    \Delta L \approx \nabla M\bigl(\bp(t_k)\bigr) \cdot \Delta \bp (t_k)
    = \pol_k C,
    \label{eq:LEGMmap}
\end{equation}
where $\Delta \bp (t_k) \doteq \bp(t_k) - \bp(t_k - \Delta t_k)$ is the map displacement ``traveled'' by the pixel $\bx_k$ as the camera moves during $\Delta t_k$.
Hence, the LEGM \eqref{eq:LEGMmap} naturally associates each event $e_k$ with the brightness gradient at one map point, $\nabla M\bigl(\bp(t_k)\bigr)$.

\subsection{Problem Formulation}
\label{sec:method:emba}

\textbf{Objective or Loss Function}.
\label{sec:method:emba:objective}
Stemming from \eqref{eq:LEGMmap}, each event represents a brightness change of predefined size $C$, 
which can be modeled in terms of the camera motion $\Rot(t)$ and the scene texture $\nabla M$. 
Hence, assuming $C$ is known, a natural design choice consists of formulating the BA problem as finding the motion and scene parameters $\bP$ that minimize the sum of square errors
\begin{equation}
\loss(\bP) \doteq \sum_{k=1}^{\numEvents} \bigl( \hat{\Delta L}_k(\bP) - \Delta L_k \bigr)^{2}
\label{eq:ObjFunc}
\end{equation}
where $\Delta L_k \doteq \pol_k C$ is the measurement, $\hat{\Delta L}_k (\bP) \doteq \nabla M \cdot \Delta \bp$ is its prediction, and $\numEvents$ is the number of events considered.
Stacking the per-event error terms into a vector 
$(\be)_k \doteq \hat{\Delta L}_k(\bP) - \Delta L_k$, we may rewrite the problem as: 
\begin{equation}
\label{eq:NLLSGenericProblem}
\min_{\bP}\loss(\bP),\quad \text{with}\quad \loss = \|\be\|^2 = \be^{\top}\be,
\end{equation}
where $\be(\bP) \in \Real^{\numEvents}$ is the photometric error (or ``residual'') vector.
This is a non-linear least squares (NLLS) function of the state $\bP$. 
It admits the probabilistic interpretation of maximum likelihood estimation under the assumption of zero-mean Gaussian noise in the temporal contrast $\Delta L$, 
which is a sensible choice according to empirical evidence \cite[Fig.6]{Lichtsteiner08ssc}.

\textbf{Solution Approach}.
\label{sec:method:emba:solution}
The standard and effective approach to minimize NLLS objectives like \eqref{eq:NLLSGenericProblem} is Gauss-Newton's (GN) method and its variations, e.g., Levenberg-Marquardt (LM) \cite{Hartley03book,Barfoot15book}. 
They linearize the errors in terms of the parameters, solve the normal equations and update the model parameters $\Delta\bP^{\ast}$, iterating until local convergence.
For GN, assuming an ``operating point'' $\bP_{\text{op}}$ (in a high dimensional space) and a perturbation $\Delta\bP$ around this operating point, the errors are linearized in terms of the parameters:
\begin{equation}
\be \approx \be_{\text{op}} + \mJ_{\text{op}}\Delta\bP,
\label{eq:ErrorsLinearized}
\end{equation}
where $\be_{\text{op}}=\be(\bP_{\text{op}})$, and $\mJ_{\text{op}}$ is the derivative of the error with respect to $\bP$.
Inserting \eqref{eq:ErrorsLinearized} into \eqref{eq:NLLSGenericProblem}, differentiating with respect to $\Delta\bP$ and setting the result equal to zero yields the necessary optimality condition.
The optimal perturbation $\Delta\bP^{\ast}$ satisfies the system
of normal equations:
\begin{equation}
\mJ_{\text{op}}^{\top}\mJ_{\text{op}} \Delta\bP^{\ast} = -\mJ_{\text{op}}^{\top}\be_{\text{op}}
\quad\Leftrightarrow\quad
\mA\, \Delta\bP^{\ast} = \bb 
\label{eq:NormalEqs}
\end{equation}
The optimal perturbation is used to update the ``operating point'' and iterate.

While this approach may appear as a classic one, several challenges are involved:
($i$) designing a meaningful and well-behaved loss, 
($ii$) identifying a suitable parametrization,
($iii$) finding efficient approximations and solvers for an actual implementation.
We tackle these challenges in the upcoming paragraphs.

\textbf{Parameterization}.
\label{sec:method:param}
The two unknowns of the problem are the camera trajectory $\Rot(t)$ and the scene map $M$. 
The continuous-time trajectory is approximated using splines that interpolate $\Rot(t)$ linearly between two neighboring poses $\{\Rot_{i},\Rot_{i+1}\} \subset \balpha$.
Thus the parameters $\balpha$ represent the discrete ``control poses'' that specify the trajectory \cite{Guo24tro}.
The map $M$ is approximated by a panoramic intensity image (\cref{fig:map_comp}).
Since the error terms $(\be)_k$ depend directly on the intensity gradient $\nabla M$, we use its $\numPixels$ pixels $\bbeta$ as the parameters to optimize in~\eqref{eq:NLLSGenericProblem}.

The computation of the linearized errors \eqref{eq:ErrorsLinearized} in terms of the camera and scene parameters $\balpha$ and $\bbeta$ is given in the \gred{supplementary}. 
We use a Lie Group sensible LM approach \cite{Barfoot15book} to linearize and update camera rotations.
The perturbation $\Delta \bP$ of the parameters has two parts, corresponding to the camera trajectory $\Delta \bP_{\balpha} \in \Real^{3 \numPoses}$ (with 3 DOFs per control pose), 
and the map pixels $\Delta \bP_{\bbeta} \in \Real^{2 \numPixels}$ (with 2 values/channels per map gradient pixel).

\begin{figure}[t]
     \centering
     \begin{subfigure}[c]{0.48\linewidth}
         \centering
         \includegraphics[width=\linewidth]{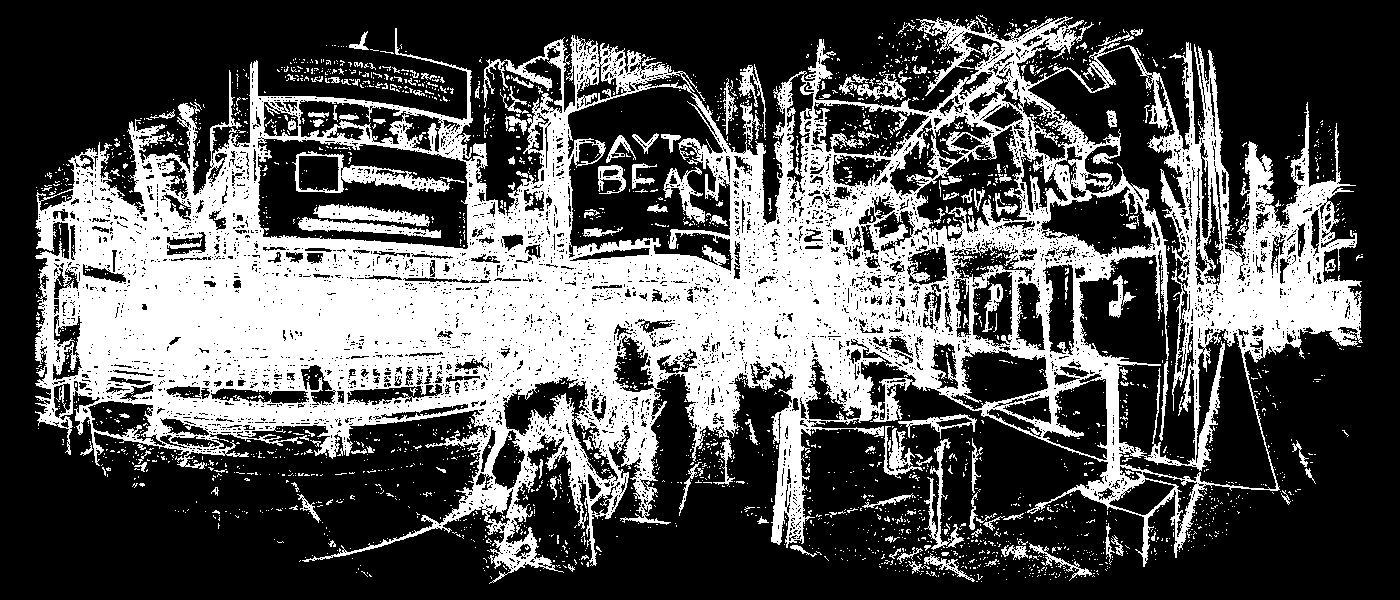}
         \caption{Mask of valid pixels.}
         \label{fig:sparsity:mask}
     \end{subfigure}
     \hspace{1ex}
     \begin{subfigure}[c]{0.195\linewidth}
         \centering
         \gframe{\includegraphics[width=\linewidth]{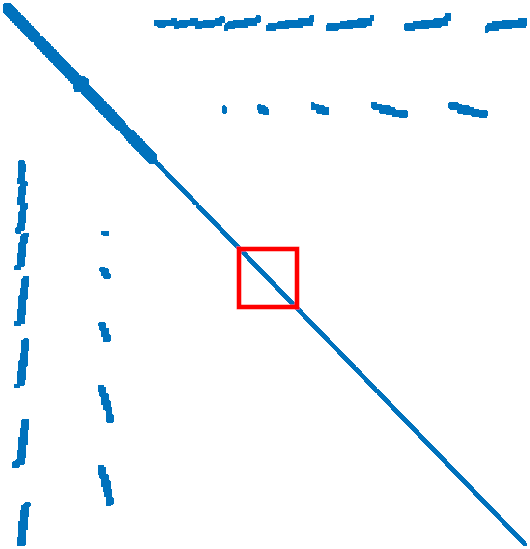}}
         \caption{Top left of $\mA$.}
         \label{fig:sparsity:A}
     \end{subfigure}
     \hspace{1ex}
     \begin{subfigure}[c]{0.20\linewidth}
         \centering
         \gframe{\includegraphics[width=\linewidth]{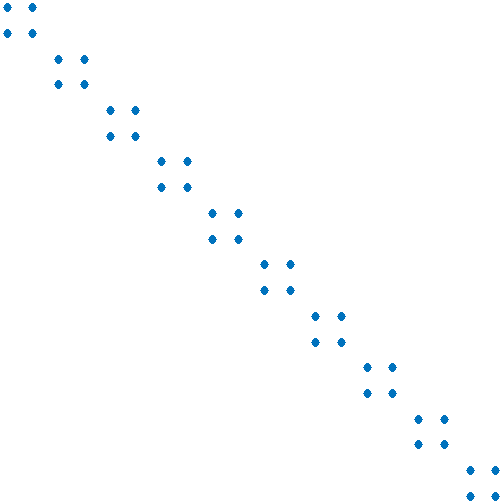}}
         \caption{Zoomed in.}
         \label{fig:sparsity:blk}
     \end{subfigure}
        \caption{\emph{Sparsity illustration}. (a) Mask of valid pixels; 
        (b) $1000 \times 1000$ block at the top left of matrix $\mA$; 
        (c) Zoomed-in version of $\mA_{22}$ showing its block-diagonal structure. 
        \label{fig:sparsity}}
\end{figure}

\textbf{Partitioning and Sparsity}.
\label{sec:method:sparsity}
For problems of moderate size, with millions of events ($\numEvents$), thousands of pixels ($\numPixels$), and hundreds of control poses ($\numPoses$), it is intractable to store the Jacobian matrix $\mJ_{\text{op}} \in \Real^{\numEvents \times (3 \numPoses + 2 \numPixels)}$.
Even in sparse format, accessing its non-zero entries is time-consuming because it does not have a simple sparsity pattern.
Instead, we directly calculate the matrix in the normal equations in an efficient way (see the \gred{supplementary}).
This matrix only depends on the number of unknowns, which is significantly smaller than the size of $\mJ_{\text{op}}$, 
and has a simpler sparsity pattern.

According to the parameterization, the state $\bP$ of the BA problem has two parts: the camera rotations and the scene map. 
This allows us to partition the perturbation vector and the normal equations \eqref{eq:NormalEqs} in blocks:
\begin{equation}
\left(\begin{array}{cc}
\mA_{11} & \mA_{12}\\
\mA^\top_{12} & \mA_{22}
\end{array}\right)\left(\begin{array}{c}
\Delta\bP_{\balpha}^{\ast}\\
\Delta\bP_{\bbeta}^{\ast}
\end{array}\right)
=\left(\begin{array}{c}
\bb_{1}\\
\bb_{2}
\end{array}\right),
\label{eq:NormalEqsFirstPartitioning}
\end{equation}
where $\mA_{11} \doteq \mJ_{\text{op},\balpha}^{\top}\mJ_{\text{op},\balpha}$ only depends on the derivatives w.r.t. the camera poses,
$\mA_{22} \doteq \mJ_{\text{op},\bbeta}^{\top}\mJ_{\text{op},\bbeta}$ only depends on the derivatives w.r.t. the scene map,
and $\mA_{12} \doteq \mJ_{\text{op},\balpha}^{\top}\mJ_{\text{op},\bbeta}$.
There is a large size difference: the size of $\mA_{11}$ (poses) is significantly smaller than that of $\mA_{22}$ (map pixels), as shown in \cref{fig:sparsity:A}.
This fact can be leveraged when using well-known tools for solving block-partitioned systems.

Additionally, we can exploit sparsity to implement an efficient LM solver for this problem, as follows.
Due to the sparsity of event data, only a portion of map points is observed.
We select sufficiently measured map points (e.g., receiving more than five events) as ``valid pixels'' in the optimization, as shown in \cref{fig:sparsity:mask}.
Furthermore, \eqref{eq:NLLSGenericProblem} states that each error term $(\be)_k$ only depends on the gradient at one map point, 
which leads to a block-diagonal structure of $\mA_{22}$ (with blocks of size $2 \times 2$), 
as depicted in \cref{fig:sparsity:blk}.
This makes $\mA_{22}$ easy to invert.
We can leverage this property, together with the block-partitioning structure of the normal equations \eqref{eq:NormalEqsFirstPartitioning} to solve them efficiently via the Schur complement \cite{Barfoot15book}.

\textbf{Map Regularization (Loss)}.
\label{sec:method:reg}
The fact that each error term $(\be)_k$ only depends on the gradient at one map pixel is beneficial for speed, but it causes instabilities: 
during optimization the values of $\bbeta$ at some pixels may grow rapidly, suppressing the update of other pixels.
To mitigate this, we add a map prior to \eqref{eq:ObjFunc}, so that map pixels evolve with regularization, 
yielding the objective:
\begin{equation}
\label{eq:NLLSRegProblem}
\min_{\{\Rot_i\},\nabla M}\|\be(\{\Rot_i\}, \nabla M)\|^2 + \eta \| \nabla M\|^2,
\end{equation}
where $\{\Rot_i\} \equiv \balpha$ are the control poses of the camera trajectory, 
and $\eta>0$ is the weight of the $\Ltwo$ regularizer $\| \nabla M\|^2 \equiv \|\bbeta\|^2$, which encourages smoothness of the estimated map.
Consequently, the normal equations of \eqref{eq:NLLSRegProblem} become:
\begin{equation}
\left(\begin{array}{ccc}
\mA_{11} & \mA_{12} & \multirow{2}{*}{\scalebox{1.25}{$\mZero$}}\\
\mA^\top_{12} & \mA_{22} + \eta \mOne & \\
\multicolumn{2}{c}{\scalebox{1.25}{$\mZero$}} & \eta \mOne
\end{array}\right)\left(\begin{array}{c}
\Delta\bP_{\balpha}^{\ast}\\
\Delta\bP_{\bbeta_1}^{\ast}\\
\Delta\bP_{\bbeta_2}^{\ast}
\end{array}\right)
=\left(\begin{array}{c}
\bb_{1}\\
\bb_{2} - \eta \nabla M_{\text{op}, \bbeta_1}\\
- \eta \nabla M_{\text{op}, \bbeta_2}
\end{array}\right),
\label{eq:NormalEqsRegPartitioning}
\end{equation}
where $\mOne$ is identity matrix, and we distinguish ``valid'' and ``invalid'' pixels using $\bbeta_1$ and $\bbeta_2$, respectively.
Eq.~\eqref{eq:NormalEqsRegPartitioning} says that ``invalid'' pixels are updated only using the $\Ltwo$ regularization, which just sets their gradients to zero.
For valid pixels, the $\Ltwo$ regularization adds a scaled identity matrix to $\mA_{22}$, which does not spoil its block-diagonal structure.
Hence, it is still cheap to solve \eqref{eq:NormalEqsRegPartitioning}.

\textbf{Poisson Reconstruction}.
\label{sec:method:poisson}
Having obtained the optimized gradient map $\bbeta$ by solving the above NLLS problem, 
we can recover the corresponding intensity map $M$ by solving the well-know Poisson's equation \cite{Agrawal06eccv,Kim14bmvc}:
$\nabla^2 M = \frac{\partial g_x}{\partial x} + \frac{\partial g_y}{\partial y},$
where $g_x$ and $g_y$ are the two channels of image $\bbeta \equiv \nabla M$.
\section{Experiments}
\label{sec:experim}

\def\figWidth{0.249\linewidth}
\begin{figure*}[t]
     \centering
     \begin{subfigure}[b]{0.40\linewidth}
         \centering
         \includegraphics[width=\linewidth]{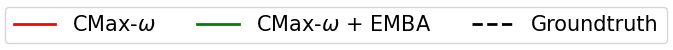}
     \end{subfigure}
     
    \begin{subfigure}[b]{\figWidth}
         \centering
         \includegraphics[width=\linewidth]{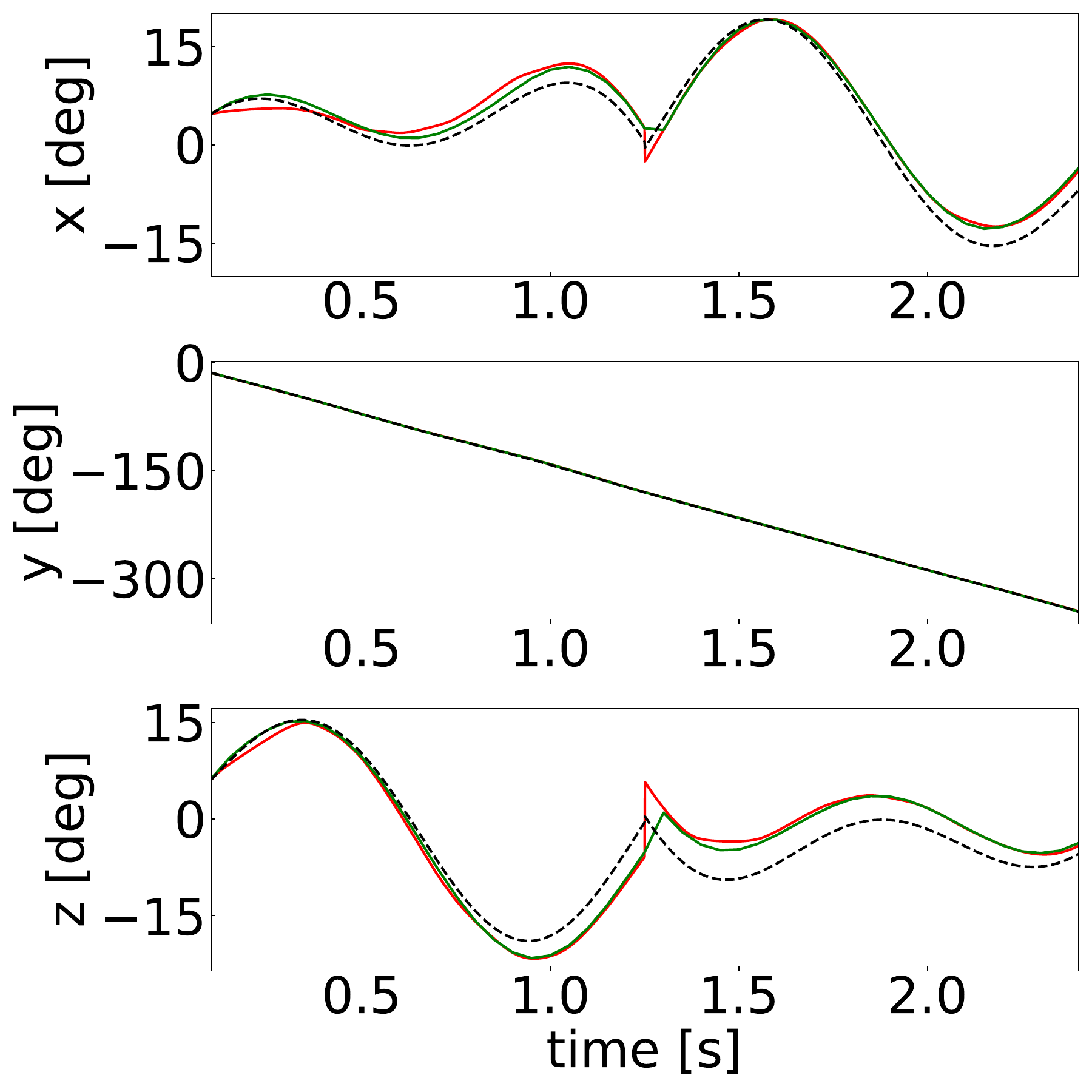}
         \caption{\playroom{} (synth)}
         \label{fig:traj:playroom}
     \end{subfigure}\,%
     \begin{subfigure}[b]{\figWidth}
         \centering
         \includegraphics[width=\linewidth]{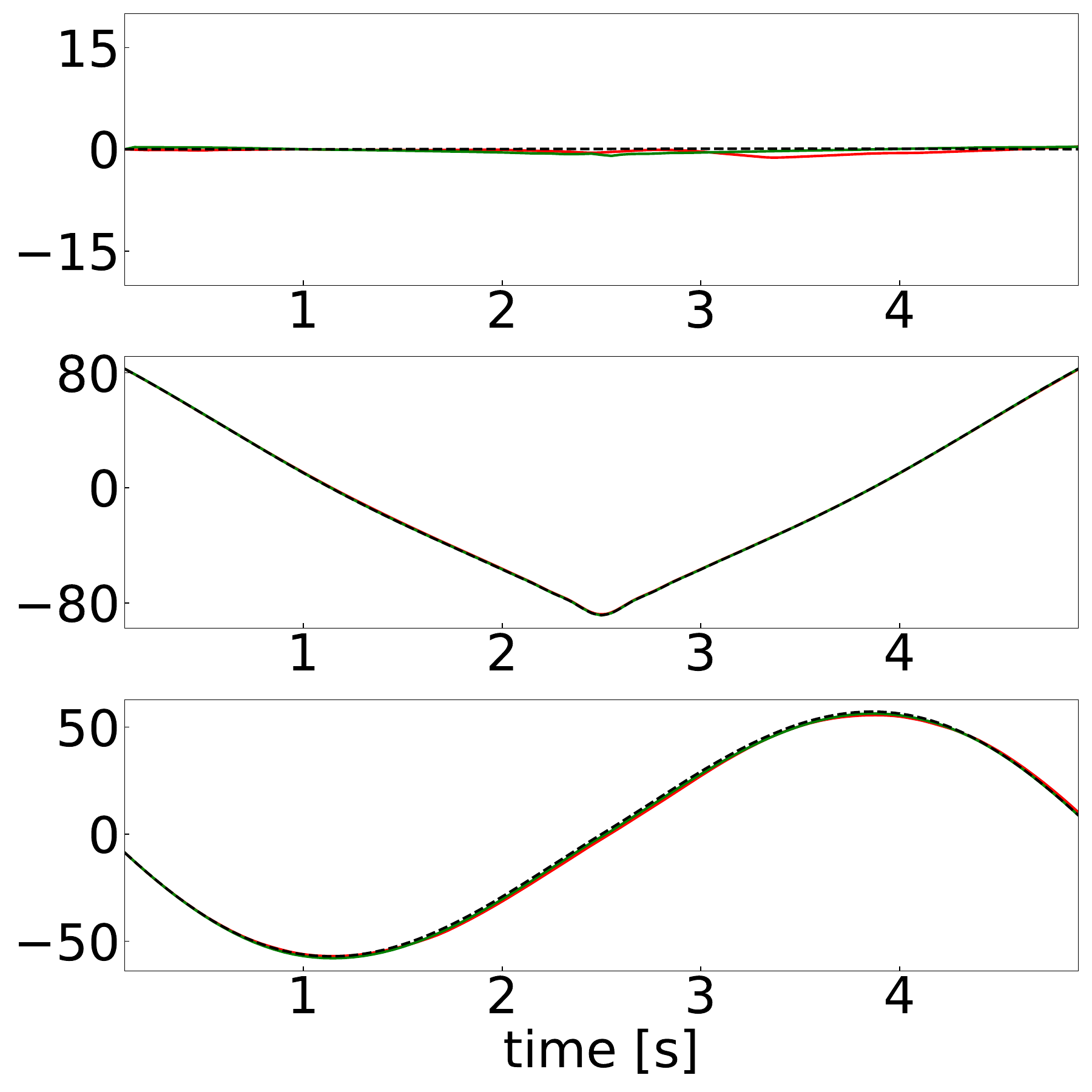}
         \caption{\city{} (synth)}
         \label{fig:traj:city}
     \end{subfigure}\,%
     \begin{subfigure}[b]{\figWidth}
         \centering
         \includegraphics[width=\linewidth]{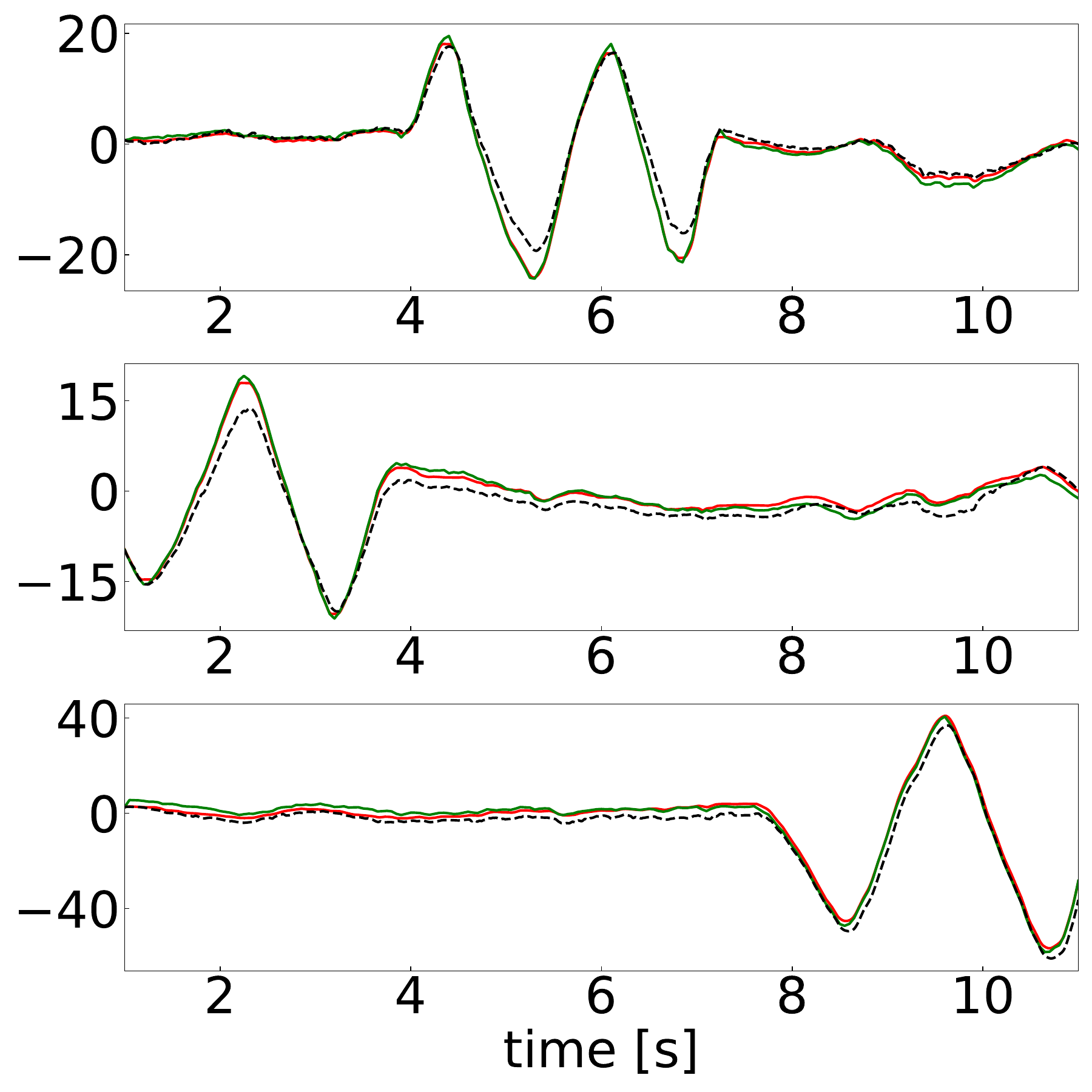}
         \caption{\shapes{} (real)}
         \label{fig:traj:shapes}
    \end{subfigure}\,%
    \begin{subfigure}[b]{\figWidth}
         \centering
         \includegraphics[width=\linewidth]{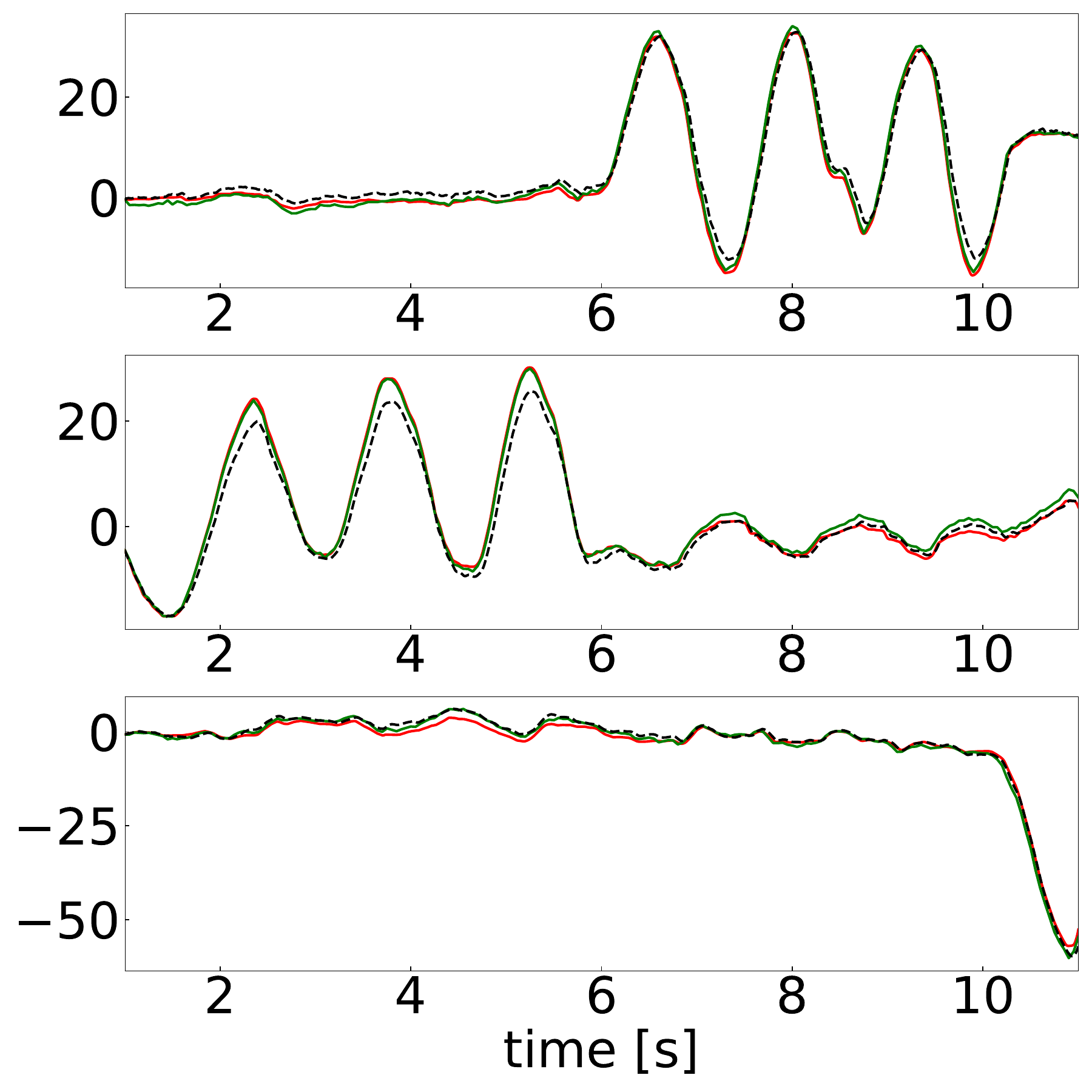}
         \caption{\dynamic{} (real)}
         \label{fig:traj:dynamic}
    \end{subfigure}%
    \caption{Camera trajectory degrees-of-freedom (DOFs) before (``\cmaxw{}'') and after (``\cmaxw{}+EMBA'') refinement, for some synthetic and real sequences from \cite{Mueggler17ijrr,Guo24tro}.}
    \label{fig:traj}
\end{figure*}

\subsection{Experimental Setup}
\label{sec:experim:setup}

\textbf{Datasets}.
We test EMBA on publicly available data: six synthetic sequences from \cite{Guo24tro} and four real-world sequences from \cite{Mueggler17ijrr}.
All sequences contain events, frames (not used), IMU data (not used) and groundtruth (GT) poses.

The synthetic sequences were obtained with a simulator \cite{Rebecq18corl}, with input panoramas from the Internet.
The panoramas covered indoor, outdoor, daylight, night, human-made and natural scenarios, 
with varying resolution, from 2K (\playroom{}), 4K (\bicycle{}), 6K (\city{} and \street{}), to 7K (\town{} and \bay{}). 
\playroom{} was created with a DVS128 camera model ($128 \times 128$ px) and a duration of 2.5s,
while the other five sequences were created with a DAVIS240C camera model (240 $\times$ 180 px) and a duration of 5~s.

The Event Camera Dataset (ECD) \cite{Mueggler17ijrr} contains four
dominantly-rotational motion sequences (\shapes{}, \poster{}, \boxes{} and \dynamic{})
that have been commonly used for benchmarking \cite{Gallego17ral,Reinbacher17iccp,Gallego18cvpr,Gallego19cvpr,Nunes21pami,Peng21pami,Gu21iccv,Kim21ral,Guo24tro}. 
They feature indoor scenes with various amounts of texture complexity and motion.
A motion capture system (mocap) outputs accurate GT poses at \mbox{200 Hz}.
We use the ECD data from 1 to 11~s for evaluation, where the camera translation is small.

\textbf{Initialization}.
\label{sec:experim:setup:initialization}
To obtain camera rotations and gradient maps to initialize EMBA, we first run the four front-end methods (\esmt{} \cite{Kim18phd}, \rtpt{} \cite{Reinbacher17iccp}, \cmaxgae{} \cite{Kim21ral} and \cmaxw{} \cite{Gallego17ral}) on all sequences.
Then we feed these front-end rotations and the event data into the mapping module of \esmt{} to obtain the initial maps (e.g., top row of \cref{fig:map_comp}).
The front-end rotations are interpolated at 1~kHz and aligned to the GT ones at $t = t_0$
($t_0 = 0.1$~s for synthetic data and $t_0 = 1$~s for real data) before they are used to obtain initial gradient maps and bootstrap EMBA.
Unless otherwise specified, the map size is set to $1024 \times 512$ px and the control pose frequency $f$ is set to 20~Hz.

\textbf{Evaluation Metrics}.
We evaluate EMBA using two metrics: 

\emph{Absolute Rotation Error (ARE)}.
The ARE measures the accuracy of the estimated camera rotations.
At timestamp $t_k$, the rotation error between the estimated rotation $\Rot_k$ and the corresponding GT rotation $\Rot_k'$ (obtained by linear interpolation),
is given by the angle of their difference $\Delta \Rot_k = \Rot_k'^\top \Rot_k$ \cite{Barfoot15book}.
Since each front-end method outputs rotations at a different rate, we calculate the errors at such timestamps and aggregate them using the Root Mean Square (RMS).
The refined rotations share the same control pose timestamps (regardless of the front-ends), hence errors are calculated on them.

\emph{Photometric Error (PhE)}.
The PhE, defined by \eqref{eq:ObjFunc}, measures the goodness of fit between the data and the estimated variables in the model, 
and is the most straightforward criterion to assess the effect of BA algorithms.

\subsection{Experiments on Synthetic Data}
\label{sec:experim:synth}
\begin{table}[t]
\centering
\caption{\label{tab:synth_traj_error}
Absolute rotation RMSE [$^\circ$] on synthetic sequences.
The best results per sequence are in bold.
``-'' means the method fails on that sequence, 
and ``\NA'' indicates that EMBA is not applicable because the corresponding front-end failed on this sequence.
\rtpt{} is not shown because it fails on all sequences.  
}
\adjustbox{max width=\linewidth}{
\setlength{\tabcolsep}{3pt}
\begin{tabular}{l*{12}{S[table-format=1.2]}}
\toprule
Sequence & \multicolumn{2}{c}{\text{playroom}}
         & \multicolumn{2}{c}{\text{bicycle}}
         & \multicolumn{2}{c}{\text{city}}
         & \multicolumn{2}{c}{\text{street}}
         & \multicolumn{2}{c}{\text{town}}
         & \multicolumn{2}{c}{\text{bay}} \\
\cmidrule(l{1mm}r{1mm}){2-3}
\cmidrule(l{1mm}r{1mm}){4-5}
\cmidrule(l{1mm}r{1mm}){6-7}
\cmidrule(l{1mm}r{1mm}){8-9}
\cmidrule(l{1mm}r{1mm}){10-11}
\cmidrule(l{1mm}r{1mm}){12-13}
& \text{before} & \text{after}
& \text{before} & \text{after}
& \text{before} & \text{after}
& \text{before} & \text{after}
& \text{before} & \text{after}
& \text{before} & \text{after} \\

\midrule

\esmt{} & 5.861 & 6.094 & 1.466 & 1.182 & 1.692 & 1.675 & 3.441 & 3.456 & 4.322 & 4.4 & 2.5 & 2.412 \\

\cmaxgae{} & 4.628 & 4.419 & 1.651 & 1.496 & \dash & \NA & \dash & \NA & 4.656 & 4.534 & \dash & \NA \\

\cmaxw{} & 3.223 & 2.856 & 1.69 & 0.923 & 1.532 & 0.973 & 0.965 & 0.744 & 1.905 & 0.858 & 1.797 & 1.409 \\

\bottomrule
\end{tabular}
}
\end{table}
\begin{table}[t]
\centering
\caption{\label{tab:synth_photo_error}
Squared photometric error [$\cdot 10^6$] on synthetic data.
}
\adjustbox{max width=\linewidth}{
\setlength{\tabcolsep}{3pt}
\begin{tabular}{l*{12}{S[table-format=1.2]}}
\toprule
Sequence & \multicolumn{2}{c}{\text{playroom}}
         & \multicolumn{2}{c}{\text{bicycle}}
         & \multicolumn{2}{c}{\text{city}}
         & \multicolumn{2}{c}{\text{street}}
         & \multicolumn{2}{c}{\text{town}}
         & \multicolumn{2}{c}{\text{bay}} \\
\cmidrule(l{1mm}r{1mm}){2-3}
\cmidrule(l{1mm}r{1mm}){4-5}
\cmidrule(l{1mm}r{1mm}){6-7}
\cmidrule(l{1mm}r{1mm}){8-9}
\cmidrule(l{1mm}r{1mm}){10-11}
\cmidrule(l{1mm}r{1mm}){12-13}
& \text{before} & \text{after}
& \text{before} & \text{after}
& \text{before} & \text{after}
& \text{before} & \text{after}
& \text{before} & \text{after}
& \text{before} & \text{after} \\

\midrule

\esmt{} & 0.347 & 0.226 & 0.515 & 0.299 & 2.623 & 2.126 & 1.821 & 1.518 & 1.88 & 1.514 & 2.259 & 1.961\\

\cmaxgae{} & 0.35 & 0.186 & 0.526 & 0.309 & \dash & \NA & \dash & \NA & 1.904 & 1.537 & \dash & \NA\\

\cmaxw{} & 0.326 & 0.151 & 0.552 & 0.295 & 2.714 & 1.978 & 1.895 & 1.336 & 1.917 & 1.425 & 2.303 & 1.827\\

\bottomrule
\end{tabular}
}
\end{table}
\def\figWidth{0.32\linewidth}
\begin{figure*}[t]
	\centering
    {\small
    \setlength{\tabcolsep}{1pt}
	\begin{tabular}{
	>{\centering\arraybackslash}m{0.3cm} 
	>{\centering\arraybackslash}m{\figWidth} 
	>{\centering\arraybackslash}m{\figWidth}
	>{\centering\arraybackslash}m{\figWidth}
        }
        \rotatebox{90}{\makecell{{\scriptsize Scene}}}
		&\includegraphics[width=\linewidth]{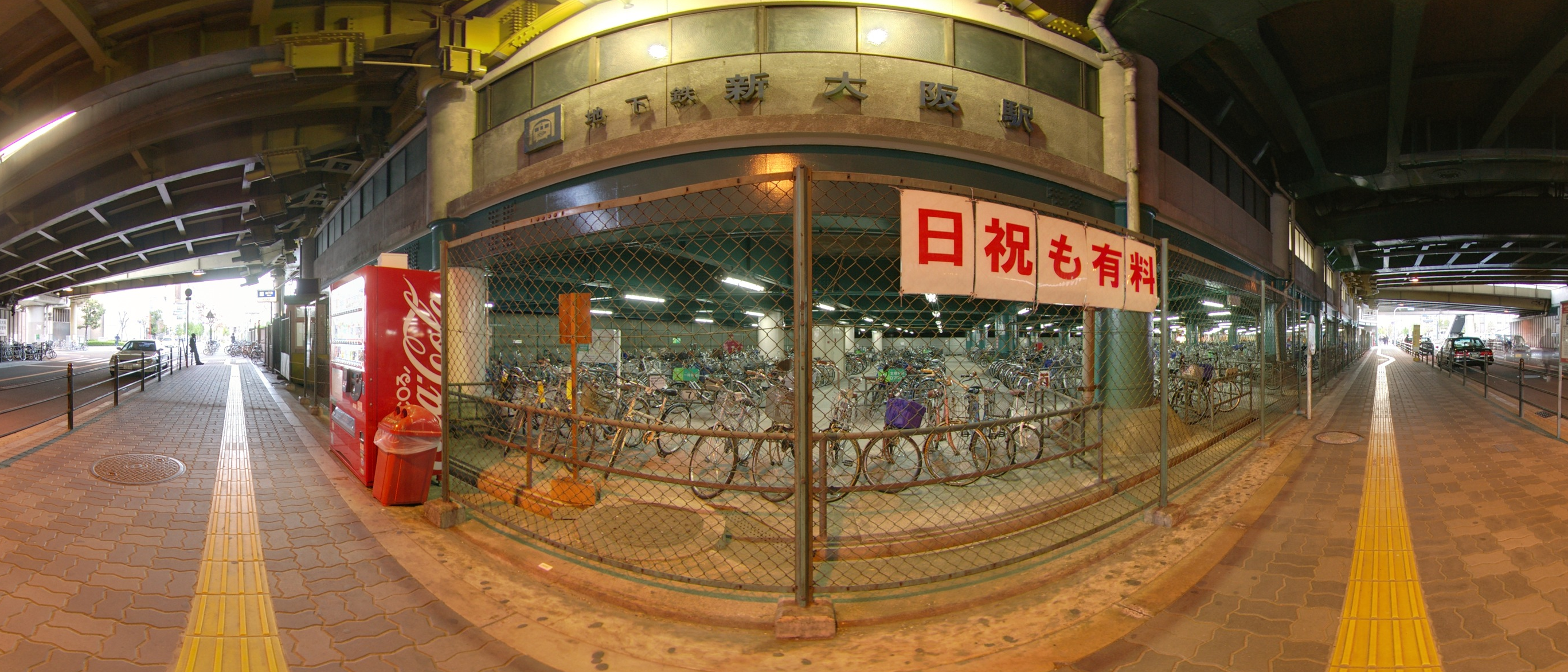}
        &\includegraphics[width=\linewidth]{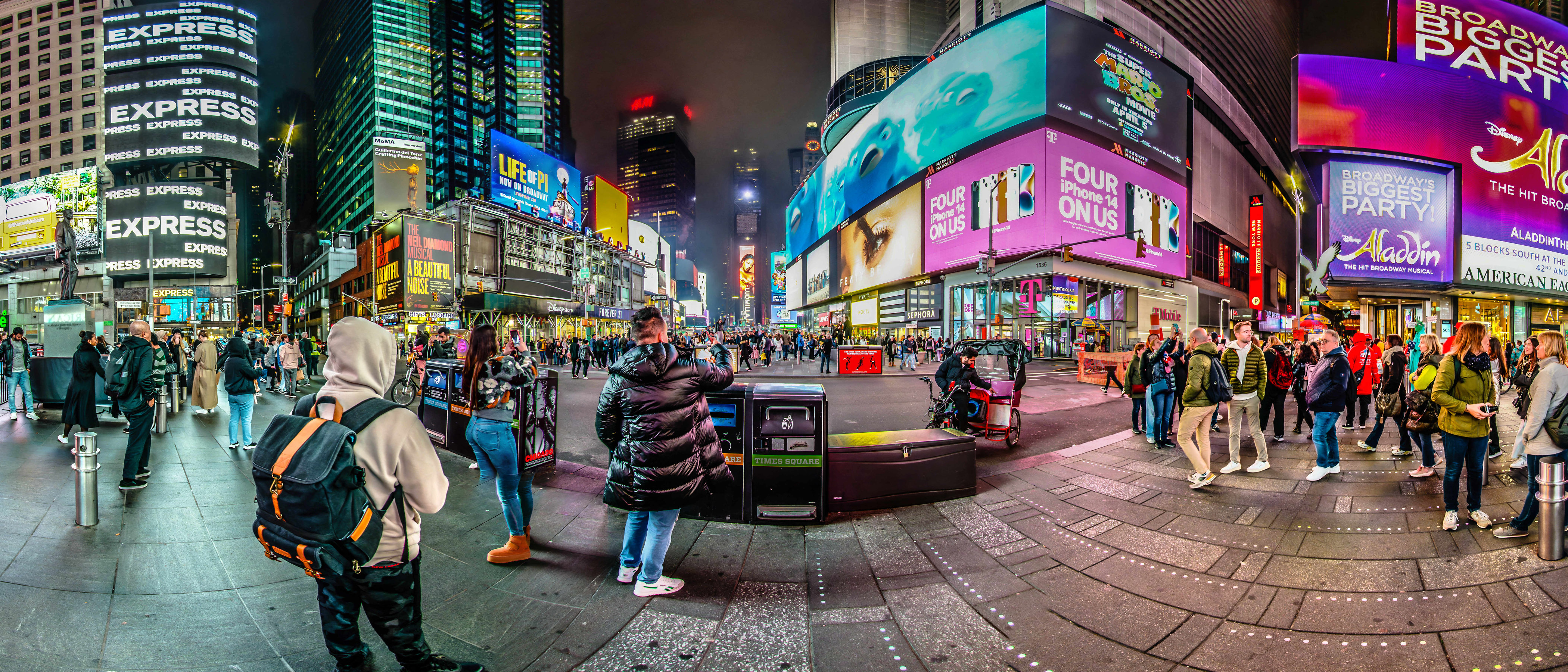}
        &\includegraphics[width=\linewidth]{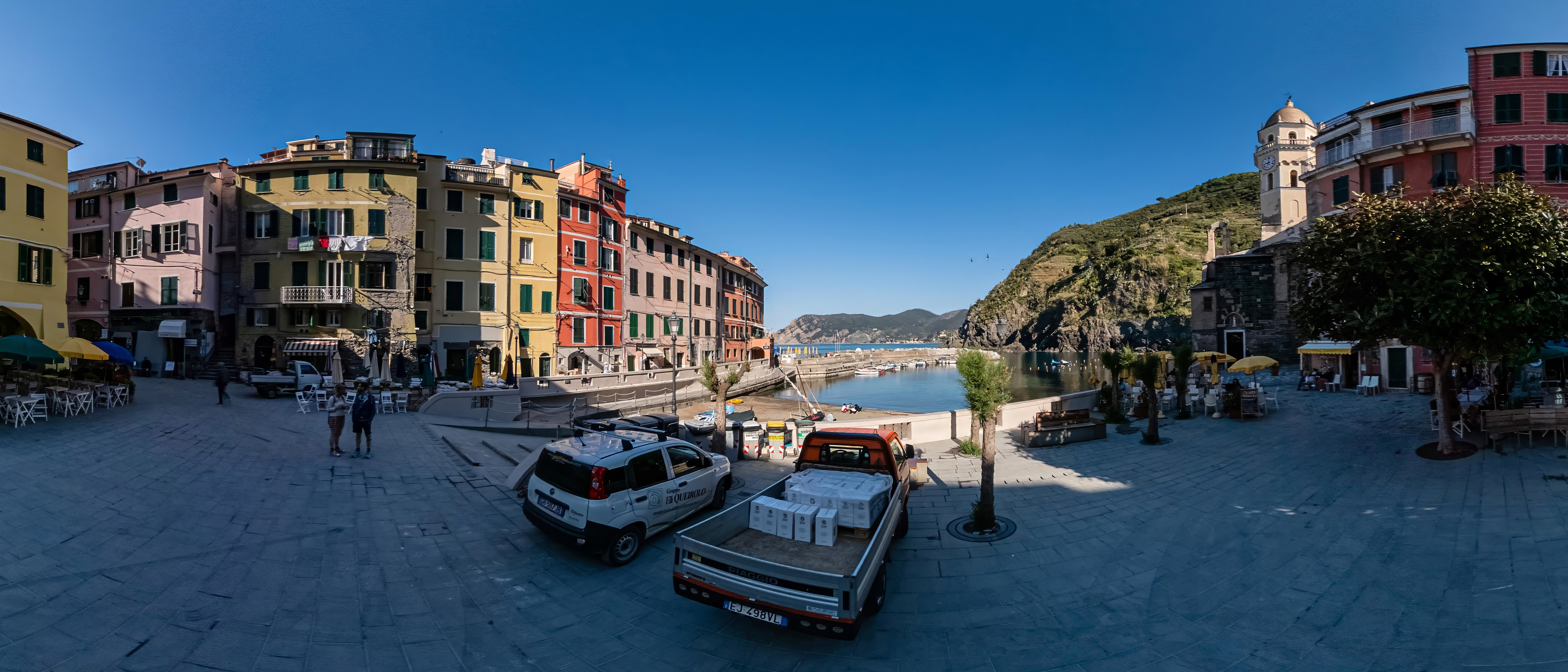}
		\\[-0.25ex]

        \rotatebox{90}{\makecell{{\scriptsize Initial map}}}
		&\includegraphics[width=\linewidth]{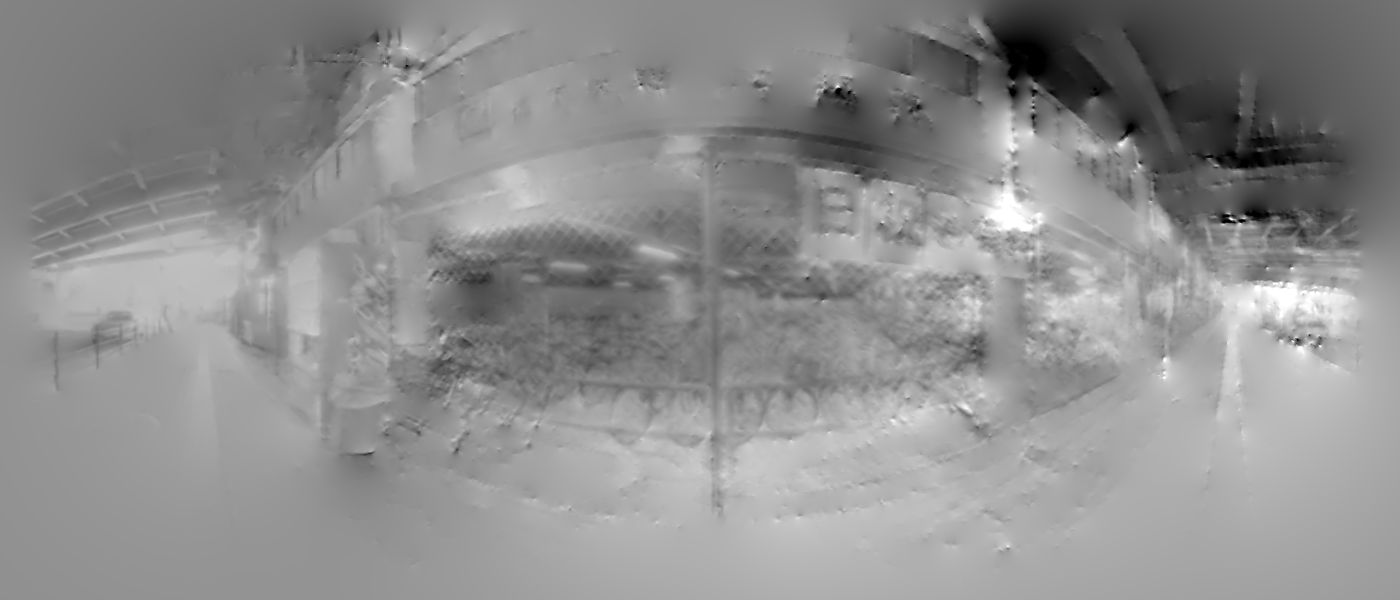}
		&\includegraphics[width=\linewidth]{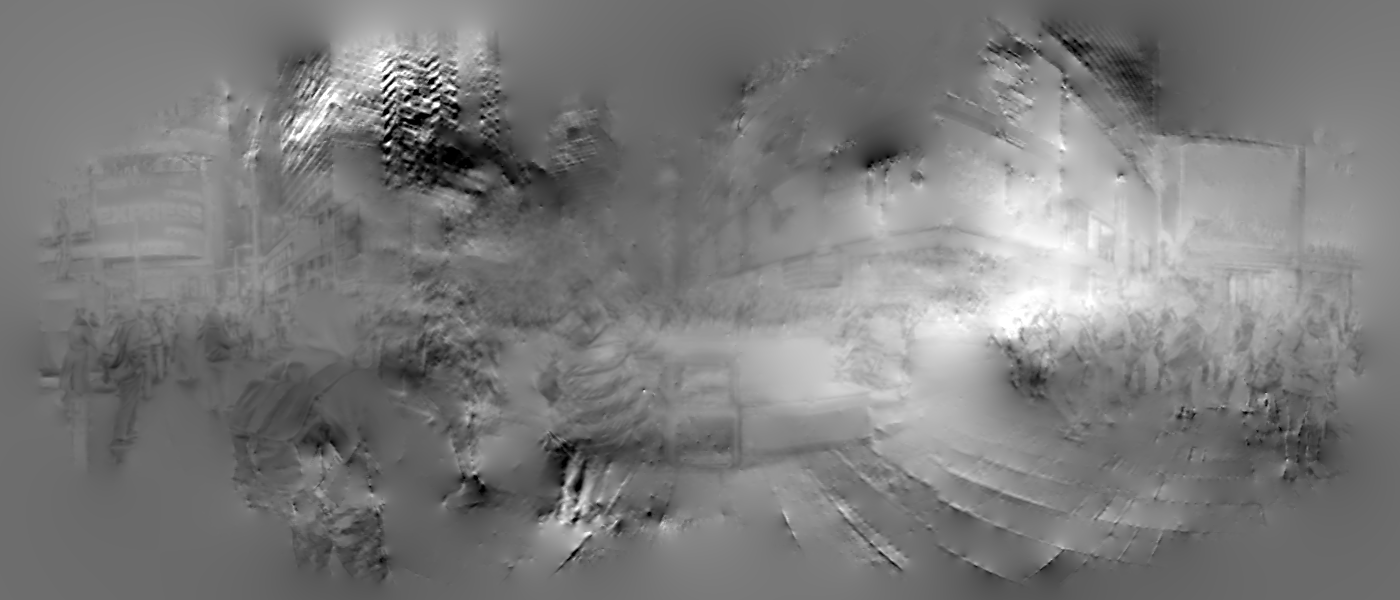}
        &\includegraphics[width=\linewidth]{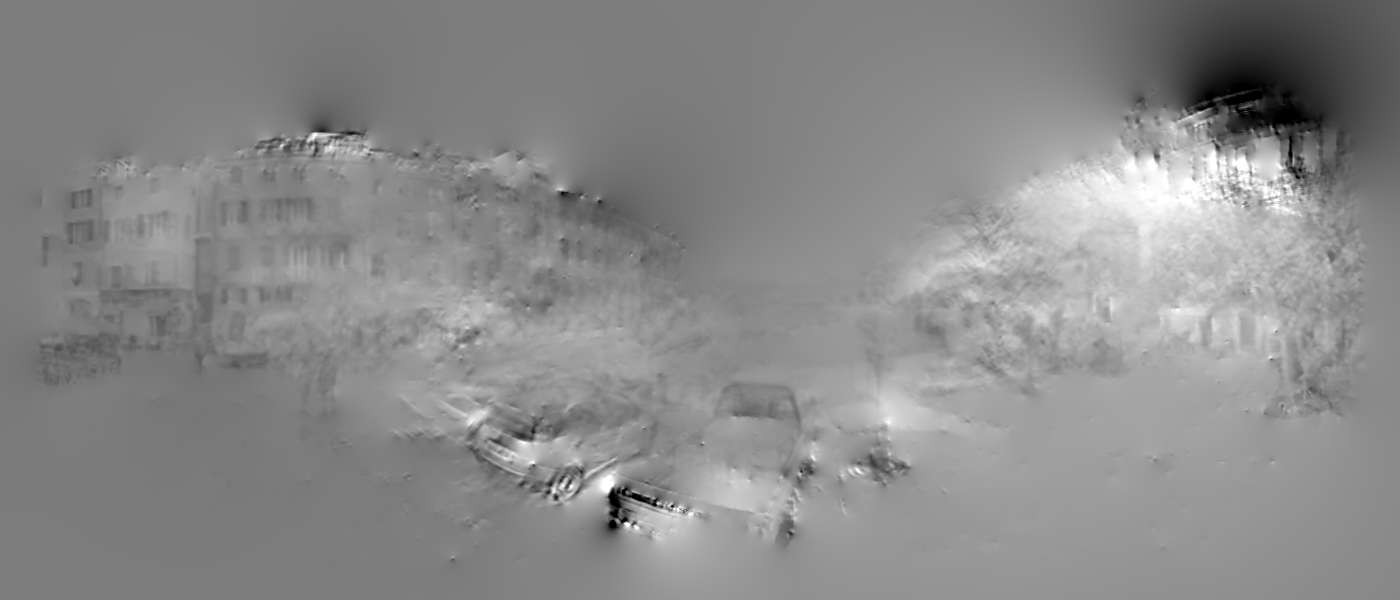}
		\\[-0.25ex]
  
        \rotatebox{90}{\makecell{{\scriptsize Refined map}}}
		&\includegraphics[width=\linewidth]{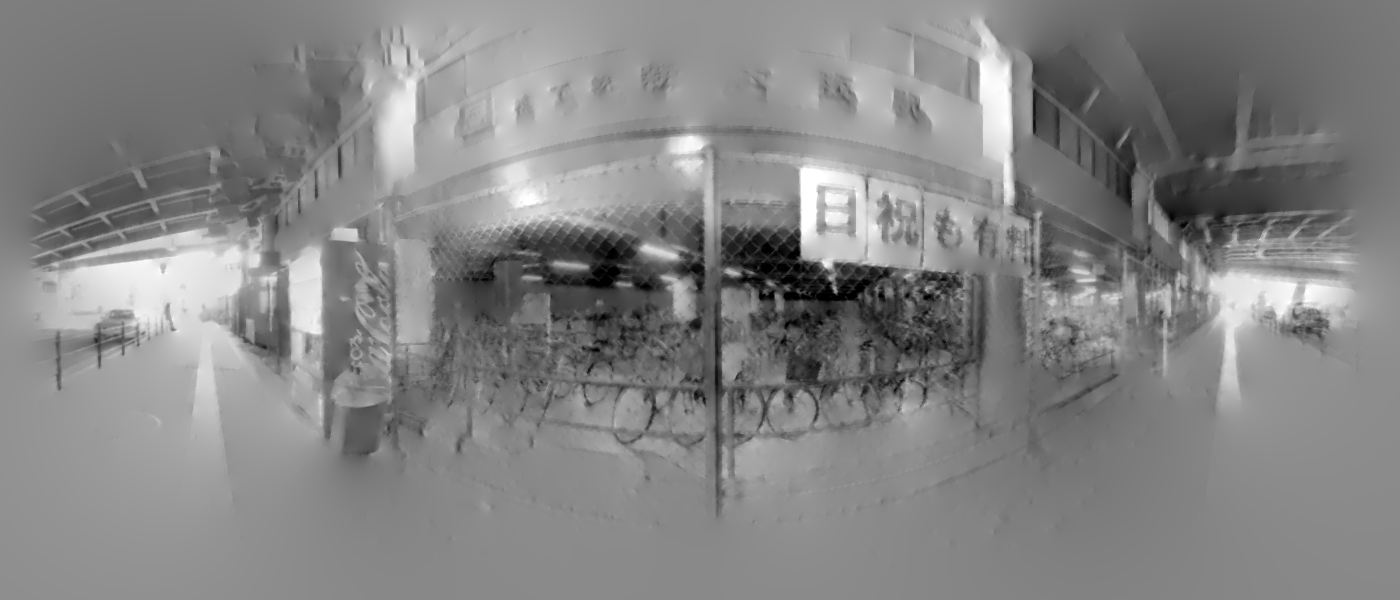}
		&\includegraphics[width=\linewidth]{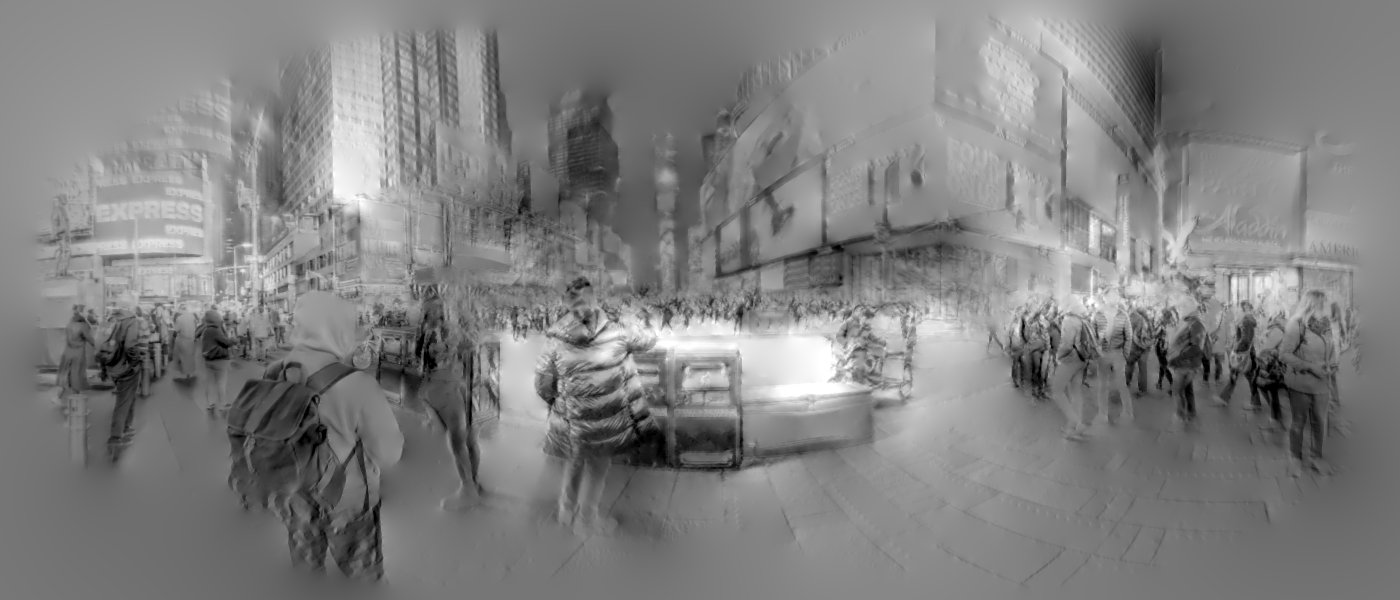}
        &\includegraphics[width=\linewidth]{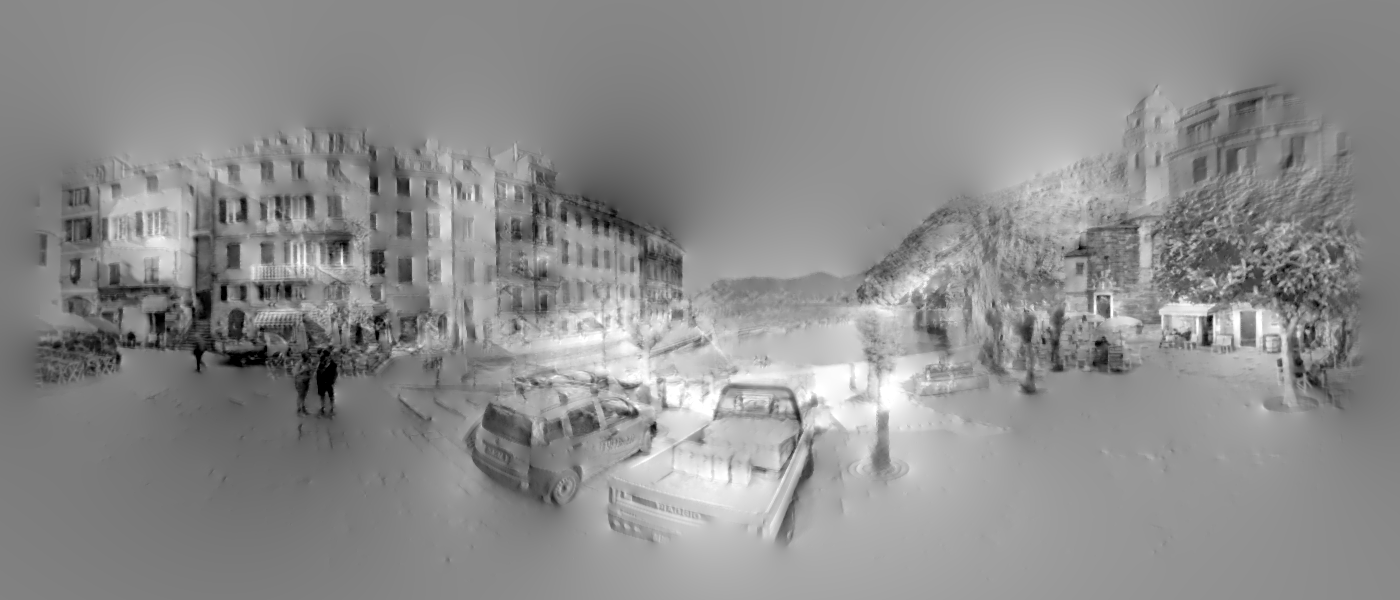}
		\\

        & (a) \bicycle{}
        & (b) \city{}
        & (c) \town{}
        \\
	\end{tabular}
	}
    \caption{\emph{EMBA results on synthetic data. }
    Panoramic maps have $2048 \times 1024$ px. 
    Initial camera rotations are obtained using \cmaxw{}~\cite{Gallego17ral}.
    \label{fig:synth_refine}
    }
\end{figure*}

\Cref{fig:traj:playroom,fig:traj:city} compare the initial and refined \cmaxw{} rotations on \playroom{} and \city{}.
The refined orientations using EMBA agree better with the GT than the initial ones.
This is further elaborated in \cref{tab:synth_traj_error}, using all front-ends: 
the errors decrease on almost all synthetic sequences, 
which is most salient when initialized by \cmaxw{}.
For \city{}, the rotation RMSE of the \cmaxw{} trajectory is reduced from 1.53$^\circ$ to 0.97$^\circ$, and that of \town{} decreases from 1.91$^\circ$ to 0.86$^\circ$.

While the plots in \cref{fig:traj} show small differences between the DOF curves, 
and the numbers in \cref{tab:synth_traj_error} also report apparently small differences (of $\approx 1^\circ$ RMSE),
the improvement effect of EMBA is most noticeable in the photometric error PhE (\cref{tab:synth_photo_error}) 
and the visual quality of the maps (\cref{fig:map_comp,fig:synth_refine}).
In all trials, the PhE values are significantly reduced (\cref{tab:synth_photo_error}); 
the maximal relative decrease reaches $54.5\%$ (when refining the \cmaxw{} rotations on \playroom{}). 

Moreover, EMBA is also able to refine higher resolution maps: 
\cref{fig:synth_refine} compares initial and refined maps of $2048 \times 1024$ px size.
The large improvements of EMBA refinement are visually obvious: 
blurred regions become sharper,
and subtle textures hidden at initialization are revealed, such as
the wheels in \bicycle{}, the billboards in \city{} and the windows and tree leaves in \town{}.

In short, EMBA achieves a compelling refinement on synthetic data in terms of 
rotation accuracy (\cref{tab:synth_traj_error}), map quality (\cref{fig:synth_refine}) and photometric error (\cref{tab:synth_photo_error}).

\subsection{Experiments on Real Data}
\label{sec:experim:real}

\begin{table}[t]
\centering
\caption{\label{tab:real_both_error}
Results on real data.
Top: absolute rotation error (ARE) [$^\circ$], in RMSE form.
Bottom: squared PhE [$\cdot 10^5$].
\esmt{} is not shown since it fails on all sequences.
}
\adjustbox{max width=.8\linewidth}{
\setlength{\tabcolsep}{4pt}
\begin{tabular}{ll*{8}{S[table-format=1.2]}}
\toprule
& Sequence & \multicolumn{2}{c}{\text{shapes}}
         & \multicolumn{2}{c}{\text{poster}}
         & \multicolumn{2}{c}{\text{boxes}}
         & \multicolumn{2}{c}{\text{dynamic}}
         \\
\cmidrule(l{1mm}r{1mm}){3-4}
\cmidrule(l{1mm}r{1mm}){5-6}
\cmidrule(l{1mm}r{1mm}){7-8}
\cmidrule(l{1mm}r{1mm}){9-10}
& & \text{before} & \text{after}
& \text{before} & \text{after}
& \text{before} & \text{after}
& \text{before} & \text{after}
\\

\midrule

\multirow{3}{*}{\begin{turn}{90}ARE\end{turn}}
&\rtpt{} & 2.187 & 2.85 & 3.802 & 3.958 & 1.743 & 2.319 & 2 & 2.285\\
 
&\cmaxgae{} & 2.512 & 2.691 & 3.625 & 4.094 & 2.018 & 2.4 & 1.698 & 2.004\\

&\cmaxw{} & 4.111 & 4.441 & 4.072 & 4.196 & 3.224 & 2.866 & 3.126 & 2.791\\

\midrule

\multirow{3}{*}{\begin{turn}{90}PhE\end{turn}}

&\rtpt{} & 0.675 & 0.374 & 4.692 & 2.583 & 4.459 & 2.304 & 3.294 & 2.239\\
 
&\cmaxgae{} & 0.609 & 0.383 & 5.03 & 3.069 & 4.522 & 2.927 & 3.164 & 2.393\\

&\cmaxw{} & 0.575 & 0.361 & 4.368 & 2.579 & 3.921 & 2.25 & 3.049 & 2.13\\

\bottomrule
\end{tabular}
}
\end{table}
\def\figWidth{0.235\linewidth}
\begin{figure*}[t]
	\centering
    {\small
    \setlength{\tabcolsep}{1pt}
	\begin{tabular}{
	>{\centering\arraybackslash}m{0.3cm} 
	>{\centering\arraybackslash}m{\figWidth} 
	>{\centering\arraybackslash}m{\figWidth}
        >{\centering\arraybackslash}m{\figWidth}
        >{\centering\arraybackslash}m{\figWidth}}

        \rotatebox{90}{\makecell{{\footnotesize GT rots. map}}}
        &\includegraphics[width=\linewidth]{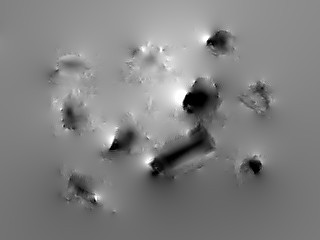}
		&\includegraphics[width=\linewidth]{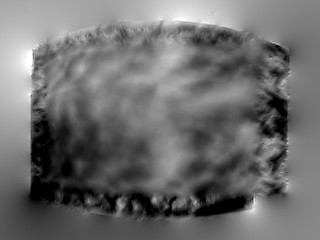}
        &\includegraphics[width=\linewidth]{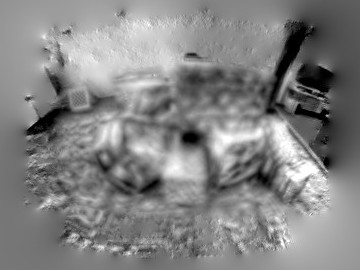}
        &\includegraphics[width=\linewidth]{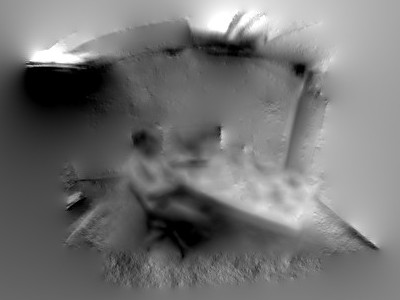}
		\\[-0.25ex]
  
        \rotatebox{90}{\makecell{{\footnotesize Initial map}}}
		&\includegraphics[width=\linewidth]{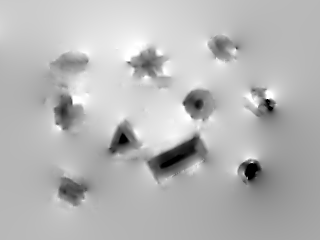}
		&\includegraphics[width=\linewidth]{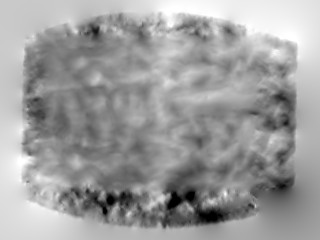}
        &\includegraphics[width=\linewidth]{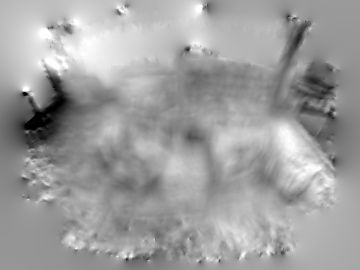}
        &\includegraphics[width=\linewidth]{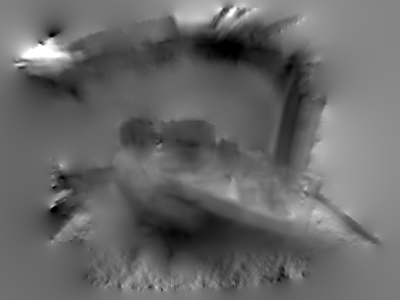}
		\\[-0.25ex]
  
        \rotatebox{90}{\makecell{{\footnotesize Refined map}}}
		&\includegraphics[width=\linewidth]{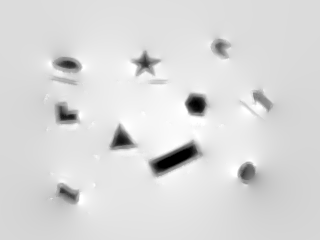}
		&\includegraphics[width=\linewidth]{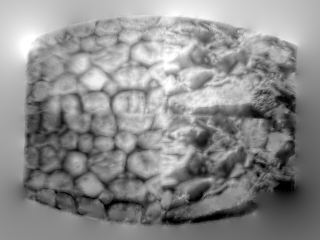}
		&\includegraphics[width=\linewidth]{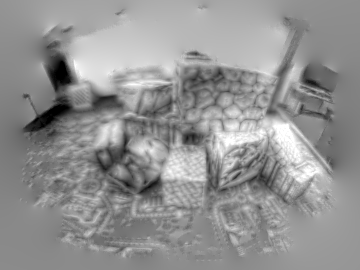}
        &\includegraphics[width=\linewidth]{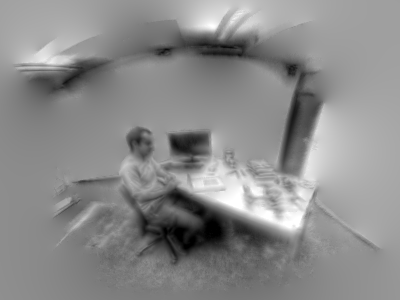}
		\\

        & (a) \shapes{}
        & (b) \poster{}
        & (c) \boxes{}
        & (d) \dynamic{}
        \\
	\end{tabular}
	}
    \caption{\emph{EMBA results on real-world data from} \cite{Mueggler17ijrr}. 
    The maps in the top two rows are obtained using the mapping module of SMT \cite{Kim14bmvc}, 
    by feeding the GT camera rotations or the rotations estimated using \cmaxw{}, respectively. 
    The refined maps are produced with our method.
    These are central crops from $1024 \times 512$ px panoramic maps.
    \label{fig:real_refine}
    }
\end{figure*}

The main difficulty of real-world evaluation lies in finding real data that is compatible with the purely rotational motion assumption of the problem \cite{Guo24tro}. 
Real-world sequences from established datasets \cite{Mueggler17ijrr} are recorded hand-held, hence they contain some translational motion (see top row of \cref{fig:real_refine}).
However, such a translation cannot be removed from the input events. 
Hence, by design, all rotational motion estimation methods explain the translation in the data using only rotational DOFs.
If the translation is non-negligible, comparing the rotations that explain additional DOFs to the rotational component of the GT provided by a 6-DOF motion-capture system \cite{Mueggler17ijrr} can be misleading.
Therefore, in the context of photometric BA, the PhE becomes a sensible figure of merit.

Plots \ref{fig:traj:shapes} and \ref{fig:traj:dynamic} compare the \cmaxw{} rotations before and after refinement on \shapes{} and \dynamic{}; the differences are small at this scale.
The top part of \cref{tab:real_both_error} reports the errors using all front-ends.
The ARE slightly decreases in some trials while it increases in others; 
there are no big differences between initial and refined trajectory errors 
because the estimated camera rotation contains compensation for the translational component.
The benefits of EMBA are demonstrated in terms of the PhE (bottom part of \cref{tab:real_both_error}) and the maps (\cref{fig:real_refine}).
EMBA considerably reduces the PhE on real-world data (\cref{tab:real_both_error}), around $30\%$ to $50\%$ reduction.
Visually, \cref{fig:real_refine} shows that a remarkable improvement is attained after refinement.
Just for comparison, we fed the GT rotations from the mocap into the mapping part of \esmt{}, and displayed the reconstructed maps in the top row in \cref{fig:real_refine}, which are blurred due to the presence of translation.
In the EMBA-refined maps (bottom row), the fine textures on the stones in \poster{} are revealed, 
and the HDR lights in the roof in \dynamic{} are also recovered clearly.

In summary, although the real-world evaluation presents difficulties, the effectiveness of EMBA is still proved by the PhE criterion and the map quality.

\subsection{Relationship with CMax-SLAM}
\label{sec:experim:cmax_slam_comp}
On the topic of event-based rotational bundle adjustment, the closest relevant work is CMax-SLAM \cite{Guo24tro}.
This section clarifies the differences between EMBA and CMax-SLAM, and demonstrates the potential of their combination.

First of all, they have different objectives and produce different types of map.
The objective of CMax-SLAM is to find the camera rotations that maximize the contrast of the panoramic IWE. 
Hence, the optimization only involves the camera rotations; 
the scene map is obtained as a secondary result and it is an edge map (\cref{fig:cmax_slam_comp:edge_map}). 
The problem is well-posed, not suffering from ``event collapse'' \cite{Shiba22sensors,Shiba22aisy,Shiba24pami}.
Conversely, EMBA aims at minimizing the event-based photometric error by refining both camera rotations and an intensity panorama.
The intensity map is explicitly modeled as a problem unknown (i.e., it is not a by-product).

In terms of mode of operation, CMax-SLAM works in a sliding-window manner, whereas EMBA processes all events in batch.
A sliding window means that rotations far away in time are not refined; hence if CMax-SLAM runs for a very long time interval, some events might align to wrong edges, which does not happen in EMBA.
Last but not least, both methods are actually complementary: 
CMax-SLAM can be used to initialize EMBA and get a clean photometric map of the scene, as shown in \cref{fig:cmax_slam_comp}.
Here, the ARE decreases from $0.470^\circ$ (see Tab.~II in \cite{Guo24tro}) to $0.377^\circ$.
Hence, EMBA can further refine the rotations from CMax-SLAM while jointly reconstructing a precise intensity panorama.

\def\figWidth{0.325\linewidth}
\begin{figure}[t]
  \centering
  \begin{subfigure}[c]{\figWidth}
         \centering
         \gframe{\includegraphics[width=\linewidth]{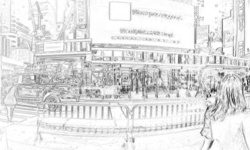}}
         \caption{Edge map \cite{Guo24tro}.}
         \label{fig:cmax_slam_comp:edge_map}
     \end{subfigure}
     \begin{subfigure}[c]{\figWidth}
         \centering
         \includegraphics[width=\linewidth]{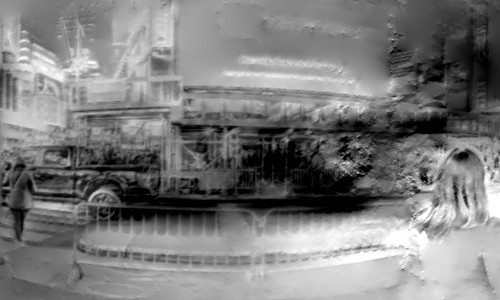}
         \caption{Initial intensity.}
         \label{fig:cmax_slam_comp:initial}
     \end{subfigure}
     \begin{subfigure}[c]{\figWidth}
         \centering
         \includegraphics[width=\linewidth]{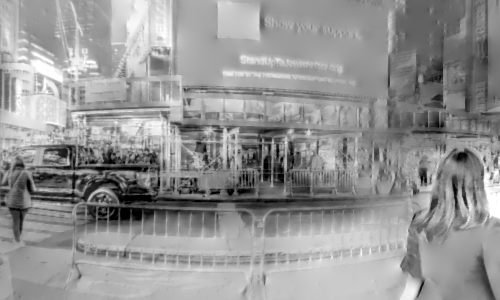}
         \caption{Refined intensity.}
         \label{fig:cmax_slam_comp:refine}
     \end{subfigure}
   \caption{Results of initializing EMBA with CMax-SLAM \cite{Guo24tro} (\emph{street} scene).}
   \label{fig:cmax_slam_comp}
\end{figure}

\subsection{Experiments with VGA and HD event cameras}
\label{sec:experim:wild}
\begin{figure}[t]
     \centering
     \begin{subfigure}[c]{0.412\linewidth}
         \centering
         \includegraphics[width=\linewidth]{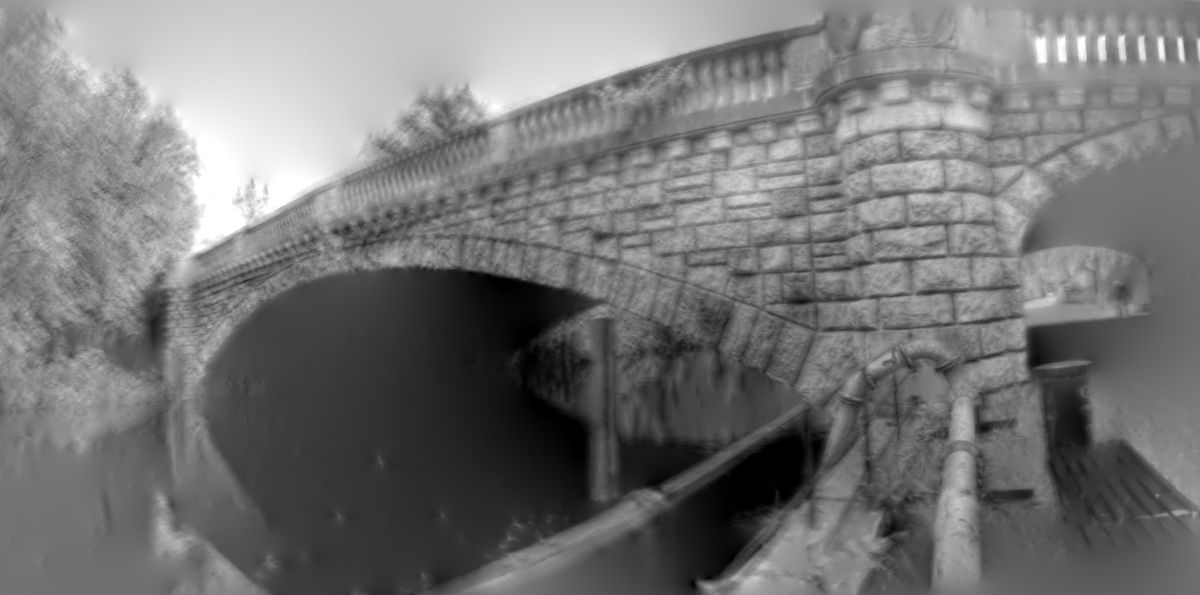}
         \caption{\bridge{}.}
         \label{fig:wild:bridge}
     \end{subfigure}\,%
     \begin{subfigure}[c]{0.5\linewidth}
         \centering
         \includegraphics[width=\linewidth]{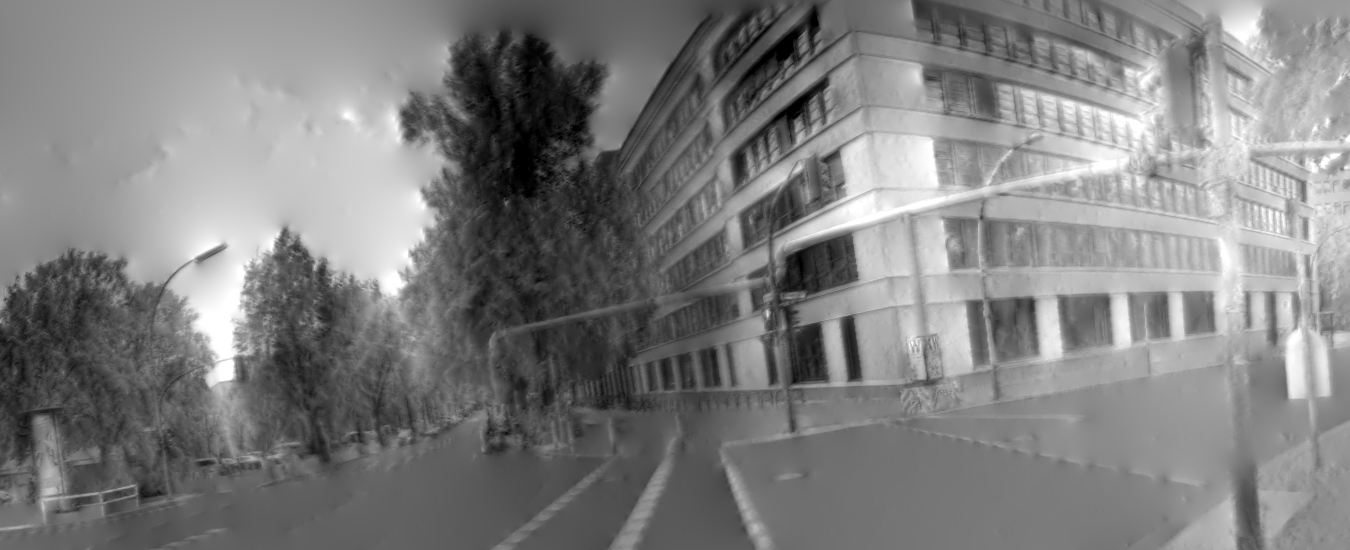}
         \caption{\crossroad{}.}
         \label{fig:wild:crossroad}
     \end{subfigure}\,%
     
     \begin{subfigure}[c]{0.455\linewidth}
         \centering
         \includegraphics[trim={2cm 0 5cm 0},clip,width=\linewidth]{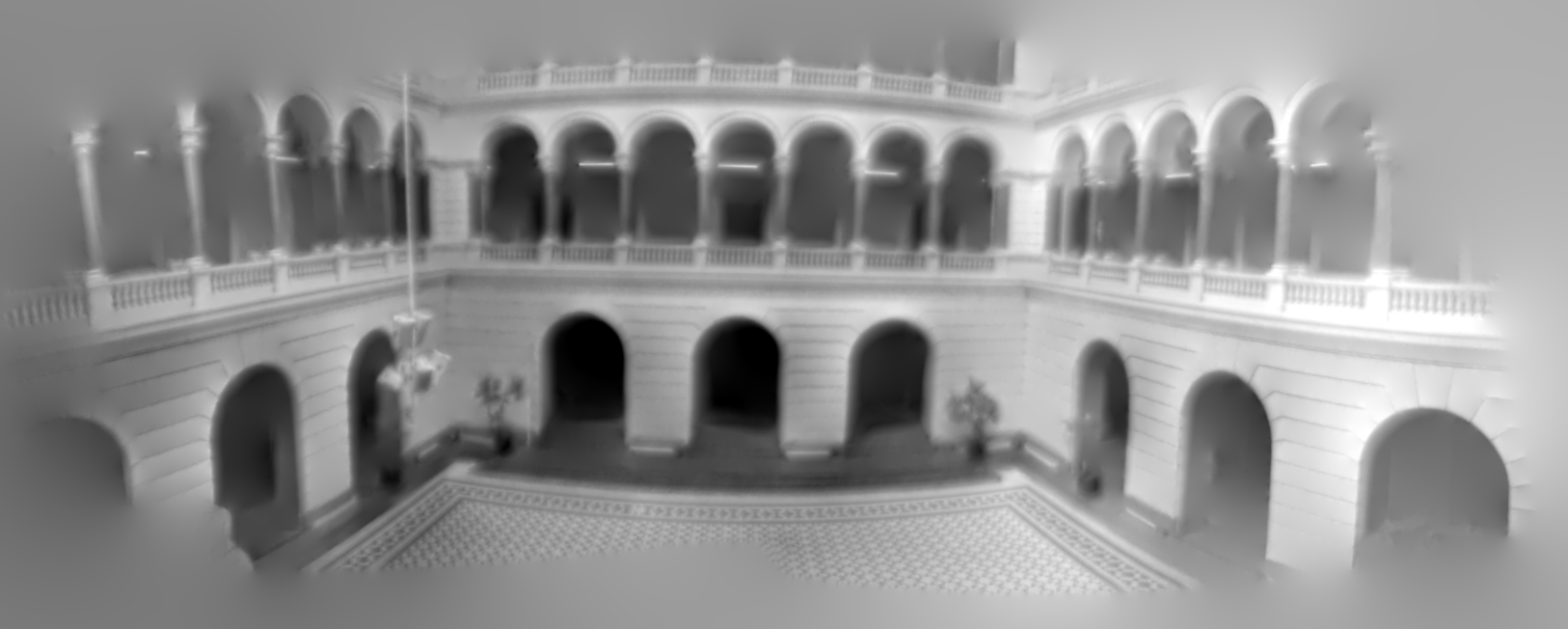}
         \caption{\atrium{}.}
         \label{fig:wild:atrium}
     \end{subfigure}\,%
     \begin{subfigure}[c]{0.455\linewidth}
         \centering
         \includegraphics[trim={0 0 0 0.9cm},clip,width=\linewidth]{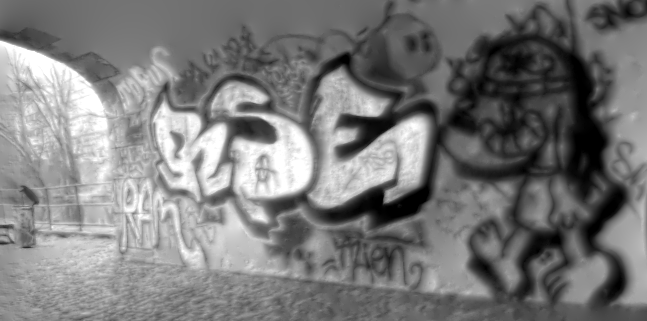}
         \caption{\graffiti{}.}
         \label{fig:wild:dummy}
     \end{subfigure}%
        \caption{\emph{Panoramas obtained from scratch.} 
        (a) and (b): DVXplorer data.
        (c) and (d): Prophesee's EVK4 (1 Mpixel camera) data.
        Crops from 4K panoramic maps.
        \label{fig:wild}}
\end{figure}

An immediate application of EMBA is panoramic imaging (mosaicing), in particular using the latest event cameras, which produce massive event rates due to their high spatial resolution. 
To this end, we show results on sequences with a DVXplorer (VGA, $640\times480$ px \cite{Son17isscc}) and a Prophesee EVK4 (HD, $1280\times720$ px \cite{Finateu20isscc}).
For the DVXplorer, which is equipped with an IMU, EMBA is initialized by IMU angular velocity dead-reckoning \cite{Barfoot15book}.
For the Prophesee EVK4, which does not have an IMU, we feed the events into \cite{Gallego17ral,Guo24tro} to provide initial camera orientations.
EMBA recovers the gradient maps from scratch while refining the camera motion parameters. 
The output panoramas in \cref{fig:wild} have high quality, demonstrating the capabilities of EMBA to handle sequences in the wild.

\subsection{Complexity Analysis and Runtime}
\label{sec:experim:runtime}
EMBA has three main steps.
First, evaluating the objective function and its derivatives, whose complexity is $O(\numEvents)$.
Second, forming the normal equations, whose complexity is also $O(\numEvents)$.
Third, solving the (LM-augmented) normal equations, whose main complexity lies in working with $\mA_{22}$.
Due to the block-diagonal structure, the cost of inverting $\mA_{22}$ 
is linear with the number of blocks, i.e., $O(\numPixels)$, where $\numPixels$ is the number of valid pixels.
Overall, EMBA is lightweight and efficient, compared to the other event-based algorithms.

To support the above statements, a runtime evaluation is carried out. 
\Cref{tab:runtime} reports the average runtime of each step for different scenes (e.g., texture complexity),
on a standard laptop (Intel Core i7-1165G7 CPU @ 2.80GHz).
The most expensive step is evaluating the objective function and its derivatives.
Obviously, the runtime of EMBA increases with $\numEvents$.
Leveraging the block-diagonal sparsity, the cost of solving the normal equations using the Schur complement increases slowly as the texture complexity grows (order: \shapes{} $<$  \dynamic{} $<$ \boxes{} $<$ \poster{}).
For comparison, we also implement EMBA with Eigen's \cite{Eigenweb} built-in conjugate gradient (CG) solver, which does not directly exploit sparsity.
We find that the Schur solver is even faster than the CG solver when $\numPixels$ is large.

\begin{table}[t]
\centering
\caption{\label{tab:runtime}Runtime evaluation of the three main steps of EMBA [s].}
\adjustbox{max width=0.9\columnwidth}{
\setlength{\tabcolsep}{8pt}
\begin{tabular}{lrrrr}
\toprule 
ECD sequence  & shapes & poster & boxes & dynamic \\
\midrule
Obj. func. evaluation & 1.114 & 8.873 & 7.436 & 5.837 \\
Forming Normal Eqs. & 0.300 & 2.366 & 2.106 & 1.574 \\
Solving Normal Eqs. (Schur) & 0.429 & 2.013 & 2.006 & 1.656 \\
Solving Normal Eqs. (CG) & 0.267 & 3.127 & 3.561 & 2.056 \\
\midrule
$N_p$ (valid pixels) & 6913 & 50738 & 49357 & 41313 \\
$N_e$ (number of events) & 1.78M & 12.59M & 10.76M & 8.80M \\
\bottomrule
\end{tabular}
}
\end{table}

\subsection{Limitations}
\label{sec:limitations}
Event cameras rely on scene texture to produce data.
Too little texture usually leads to tracking failure (EMBA initialization failure), 
while high texture triggers too many events, which slows down the algorithm in spite of the linear complexity of EMBA. 
This limitation, shared by most event-based algorithms, might be overcome by adapting the camera's $C$ value and/or downsampling events.

All surveyed event-based rotational SLAM methods assume static scenarios and brightness constancy.
Events triggered by moving objects and flickering lights may cause inaccuracy or failure if they are plentiful.
Also, the linearization in \eqref{eq:LEGM} due to Taylor's approximation in the LEGM model \cite{Gallego20pami}
is another source of inaccuracies.
However, this was chosen because it endowed matrix $\mA_{22}$ in the normal equations with a highly beneficial block-diagonal sparsity pattern.

The \LM{} method has its limitations, e.g., local convergence. 
EMBA may get stuck in local minima of the very large search space if the initialization is not good.
This is also a problem in BA for frame-based cameras.

\section{Conclusion}
\label{sec:conclusion}
We have introduced EMBA, an event-only mosaicing bundle adjustment approach to jointly refine the orientations of a rotating camera and the panoramic gradient map of the scene.
We have leveraged the LEGM to formulate the BA problem as a regularized NLLS optimization with a beneficial sparsity pattern so that it can be efficiently solved by exploiting well-developed tools in BA, such as the LM method.
To the best of our knowledge, no previous work has constructed and utilized such a useful sparsity for event-based BA without converting events into image-like representations. 
Through a comprehensive evaluation, the proposed method achieves remarkable results in terms of photometric error (50\% decrease), camera poses and map quality.
In addition, we have demonstrated the application of EMBA to mosaicing with high-resolution event cameras, of relevance for smartphone applications, even without map initialization.
We release the code and hope that our work helps bring maturity to event-based SLAM and related applications.

\section*{Acknowledgments}
Funded by the Deutsche Forschungsgemeinschaft (DFG, German Research Foundation) under Germany’s Excellence Strategy – EXC 2002/1 ``Science of Intelligence'' – project number 390523135.

\ifarxiv
\appendix
\section{Supplementary Material}

\subsection{Video}
The accompanying video shows the evolution (iterations) of the proposed event-only bundle adjustment method on multiple sequences (both synthetic and real).

\subsection{Problem unknowns, Operating Point and Perturbation}
The unknowns of the problem are the camera trajectory $\Rot(t)$ and the gradient map of the scene $\boldsymbol{G} \doteq \nabla M$.
According to the chosen parameterization (\cref{sec:method:emba}), the perturbations of the camera pose at time $t$ (not necessarily a control pose) and the gradient map are:
\begin{align}
\Rot(t) & =\exp(\delta\boldsymbol{\varphi}^{\wedge})\Rot_{\text{op}}(t),\\
\boldsymbol{G} & = \boldsymbol{G}_{\text{op}}+\Delta \boldsymbol{G},
\end{align}
where we use the exponential map (notation from \cite{Barfoot15book}).
The ``operating point'' (abbreviated ``op'') consists
of the current camera trajectory (parameterized by $\numPoses$ control poses) and
the map (e.g., gradient brightness values):
\begin{equation}
\bP_{\text{op}}=\{\Rot_{1}^{\text{op}},\ldots,\Rot_{\numPoses}^{\text{op}},\bbeta_{1}^{\text{op}},\ldots,\bbeta^{\text{op}}_{\numPixels} \}.
\label{eq:OperPointCamAndMap}
\end{equation}
To linearize the errors for the Gauss-Newton / Levenberg-Marquardt algorithm, we consider pose perturbations in the Lie-group sense (control poses in the Lie group and perturbations in the Lie algebra \cite{Barfoot15book}), and pixel perturbations in gradient brightness space. 
That is, camera control poses and map pixels are perturbed according to
\begin{eqnarray}
\Rot_{i} &=& \exp(\rotperturb_{i}^{\wedge})\Rot_{i}^{\text{op}}, \label{eq:PerturbCamPoses}\\
\bbeta_n &=& \bbeta_n^{\text{op}} + \delta\bbeta_n.\label{eq:PerturbMapbeta}
\end{eqnarray}

\subsection{Linearization of Error Terms (Analytical Derivatives)}
\label{sec:suppl:LinErrTerm}
Perturbing the camera motion and the scene map we aim to arrive at
an expression like:
\begin{equation}
\be \approx \be_{\text{op}} + \mJ_{\text{op},\balpha}\Delta\bP_{\balpha}+\mJ_{\text{op},\bbeta}\Delta\bP_{\bbeta},
\label{eq:LinearizedErrorPartitionedHigh}
\end{equation}
where $\mJ_{\text{op},\balpha}\doteq\left.\prtl{\be}{\bP_{\balpha}}\right|_{\text{op}}$
and $\mJ_{\text{op},\bbeta}\doteq\left.\prtl{\be}{\bP_{\bbeta}}\right|_{\text{op}}$.
Thus, we only consider the first-order terms (i.e., discard higher
order ones). Here, $\mJ_{\text{op},\balpha}$ is an $N_{e}\times 3\numPoses$
matrix, and $\mJ_{\text{op},\bbeta}$ is an $N_{e} \times 2\numPixels$ matrix,
where $N_{e}$ is the number of events and $\numPixels$ is the number
of valid panorama pixels.

Let us write the linearization of each error term in \eqref{eq:LinearizedErrorPartitionedHigh}.
Given the error entry from the problem \eqref{eq:ObjFunc}-\eqref{eq:NLLSGenericProblem}:
\begin{equation}
    (\be)_k \doteq \boldsymbol{G}\bigl(\bp(t_k)\bigr) \cdot \Delta \bp (t_k) - \pol_k C.
\end{equation}
After some calculations, we have:
\begin{align}
(\be)_k & \approx\left(\boldsymbol{G}(\bp_{\text{op}}(t_{k}))-\nabla\boldsymbol{G}_{\text{op}}(\bp_{\text{op}}(t_{k}))\mE_{\text{op}}(t_{k})\delta\boldsymbol{\varphi}(t_{k})+\Delta\boldsymbol{G}(\bp_{\text{op}}(t_{k}))\right) \nonumber\\
 & \qquad\cdot\left(\Delta\bp_{\text{op}}-\left(\mE_{\text{op}}(t_{k})\delta\boldsymbol{\varphi}(t_{k})-\mE_{\text{op}}(t_{k}-\Delta t_{k})\delta\boldsymbol{\varphi}(t_{k}-\Delta t_{k})\right)\right) \nonumber\\
 & \qquad -\pol_k C \\
 & \approx\underbrace{\boldsymbol{G}(\bp_{\text{op}}(t_{k}))\cdot\Delta\bp_{\text{op}}-\pol_k C}_{\text{this is } (\be_{\text{op}})_k}+\underbrace{\Delta\bp_{\text{op}}^{\top}\Delta\boldsymbol{G}(\bp_{\text{op}}(t_{k}))}_{\text{linear in }\Delta\bP_{\bbeta}} \nonumber\\
 & \qquad \underbrace{-\Delta\bp_{\text{op}}^{\top}\nabla\boldsymbol{G}_{\text{op}}(\bp_{\text{op}}(t_{k}))\mE_{\text{op}}(t_{k})\delta\boldsymbol{\varphi}(t_{k})}_{\text{linear in }\Delta\bP_{\balpha}}\nonumber\\
 & \qquad\underbrace{-\boldsymbol{G}(\bp_{\text{op}}(t_{k}))\cdot\left(\mE_{\text{op}}(t_{k})\delta\boldsymbol{\varphi}(t_{k})-\mE_{\text{op}}(t_{k}-\Delta t_{k})\delta\boldsymbol{\varphi}(t_{k}-\Delta t_{k})\right)}_{\text{linear in }\Delta\bP_{\balpha}},
 \label{eq:LinErrorLEGM}
\end{align}
where
\begin{align}
    \Delta \bp_{\text{op}}(t_{k}) & \doteq \bp_{\text{op}}(t_{k}) - \bp_{\text{op}}(t_{k-1}) \\
    \mE_{\text{op}}(t) & \doteq \left.\prtl{\pi}{\bz}\right|_{\bz_{\text{op}}}\bz_{\text{op}}^{\wedge} \\
    \pi & \;\;\text{is the equirectangular projection } \mathbb{R}^3\to\mathbb{R}^2\\
    \bz(t) & = \Rot(t)\Kint^{-1}\bx^{h}\\
    \bz_{\text{op}}(t) & \doteq \Rot^{\text{op}}(t)\Kint^{-1}\bx^{h}\\
    \bx^{h} &= (x,y,1)^{\top} \text{are the homogeneous coordinates of point } \bx\\
    ^{\wedge} & \;\;\text{is the hat (skew-symmetric) operator \cite{Barfoot15book}} \\
    \delta\boldsymbol{\varphi} & \;\;\text{is the perturbation of } \Rot(t_k)\\
    \delta\tilde{\boldsymbol{\varphi}} & \;\;\text{is the perturbation of } \Rot(t_k-\Delta t_{k}) \\
    \nabla\boldsymbol{G} & \doteq \nabla^{2} M_{\text{op}} \;\;\text{is the second-order spatial derivative of } M_{\text{op}}
\end{align}
Note that $\delta\tilde{\boldsymbol{\varphi}}$ will use the two control poses closest to time $t_{k}-\Delta t_{k}$, 
which may not necessarily be the same ones as those of $\delta\boldsymbol{\varphi}$ (at time $t_{k}$). 

In therms of the problem unknowns, equation \eqref{eq:LinErrorLEGM} states that the predicted (linearized) contrast in \eqref{eq:LEGMmap} depends on: 
the event camera orientations at two different times $\{ t_{k}$, $t_{k}-\Delta t_{k} \}$ and the first two spatial derivatives of brightness at one pixel location $\bp(t_{k})$.

\subsection{Cumulative Formation of the Normal Equations}
A key step of the Levenberg-Marquardt (LM) solver is forming the normal equations.
Regarding EMBA, the size of the full Jacobian matrix $\mJ_{\text{op}}$ in \eqref{eq:ErrorsLinearized} is $\numEvents \times (3\numPoses + 2\numPixels)$.
In general, an event data sequence has millions of events, while $\numPixels$ is usually in the order of thousands.
Hence, the memory needed to compute and store $\mJ_{\text{op}}$ is unaffordable for normal PCs.
To this end, we avoid computing and storing the full $\mJ_{\text{op}}$. 
Instead, we directly compute the left-hand side (LHS) matrix $\mA \doteq \mJ_{\text{op}}^{\top} \mJ_{\text{op}}$ and the right-hand side (RHS) vector $\bb \doteq -\mJ_{\text{op}}^{\top} \be_{\text{op}}$, in a cumulative manner.

\subsubsection{LHS Matrix $\mA$}
Let $\br_k^{\top}$ be the $k$-th row of $\mJ_{\text{op}}$, which stores the derivatives of an error term $(\be)_k$.
With the partitioning in \eqref{eq:NormalEqsFirstPartitioning}, we can further write $\br_k^{\top} = (\br_{k,\balpha}^{\top}, \br_{k,\bbeta}^{\top})$,
where $\br_{k,\balpha}$ and $\br_{k,\bbeta}$ are the camera pose part and map part of $\br_k$, respectively.
Then we can rewrite the LHS matrix as the sum of the outer product of each row:
\begin{equation}
    \mA \doteq \mJ_{\text{op}}^{\top}\mJ_{\text{op}} = \sum_{k=1}^{\numEvents} \br_k \br_k^{\top} 
    = \sum_{k=1}^{\numEvents}
    \begin{pmatrix}
        {\br_{k,\balpha}} {\br_{k,\balpha}}^{\top} & {\br_{k,\balpha}} {\br_{k,\bbeta}}^{\top} \\
        {\br_{k,\bbeta}} {\br_{k,\balpha}}^{\top} & {\br_{k,\bbeta}} {\br_{k,\bbeta}}^{\top}
    \end{pmatrix}.
    \label{eq:firstRowOuterProduct}
\end{equation}
Let ${\mA_{11}}_k \doteq {\br_{k,\balpha}} {\br_{k,\balpha}}^{\top}$,
${\mA_{12}}_k \doteq {\br_{k,\balpha}} {\br_{k,\bbeta}}^{\top}$,
and ${\mA_{22}}_k \doteq {\br_{k,\bbeta}} {\br_{k,\bbeta}}^{\top}$.
They are the contributions of $(\be)_k$ to the LHS matrix $\mA$.
Then \eqref{eq:firstRowOuterProduct} becomes:
\begin{equation}
    \mA = \sum_{k=1}^{\numEvents} \mA_k = 
    \sum_{k=1}^{\numEvents}
    \begin{pmatrix}
    {\mA_{11}}_k & {\mA_{12}}_k \\
    {\mA_{12}}_k^{\top} & {\mA_{22}}_k
    \end{pmatrix}.
    \label{eq:LHSMatCumulative}
\end{equation}
It shows that the contribution of each event to $\mA$ is additive,
which offers a cumulative way to form the LHS matrix $\mA$.
As mentioned at the end of \cref{sec:suppl:LinErrTerm}, an error term depends on map gradients at one map point (nearest neighbor).
This leads to a block-diagonal sparsity pattern of ${\mA_{22}}_k$,
which significantly speeds up solving the normal equations.

\subsubsection{RHS Vector $\bb$}
\label{sec:suppl:LM_details:RHS}
Similarly, let $\bc_n$ be the $n$-th column of $\mJ_{\text{op}}$.
With the partitioning in \eqref{eq:NormalEqsFirstPartitioning}, we can rewrite $\mJ_{\text{op}}$ as 
\begin{equation}
    \begingroup
    \setlength\arraycolsep{2pt}
    \mJ_{\text{op}} = 
    \begin{pmatrix}
        \bc_{1,\balpha}, & \ldots, & \bc_{3 \numPoses,\balpha}, 
        & \bc_{1,\bbeta}, & \ldots, & \bc_{2 \numPixels,\bbeta}
    \end{pmatrix},
    \label{eq:JopColumns}
    \endgroup
\end{equation}
where $\bc_{i,\balpha} = \left.\frac{\partial \be}{\partial {\bP_{i,\balpha}}}\right|_{\text{op}}$ 
and $\bc_{j,\bbeta} = \left.\frac{\partial \be}{\partial {\bP_{j,\bbeta}}}\right|_{\text{op}}$ 
store the derivatives of the whole error vector $\be$ with respect to each component of the pose/map state.
Substituting \eqref{eq:JopColumns} into the RHS of \eqref{eq:NormalEqs}, we obtain the cumulative formula of each entry of $\bb$:
\begin{equation}
    \begin{aligned}
        {\bb_1}_i & = -\bc^\top_{i,\balpha} \be_{\text{op}} = -\sum_{k=1}^{\numEvents} \left.\frac{\partial (\be)_k}{\partial {\bP_{i,\balpha}}}\right|_{\text{op}} (\be_\text{op})_{k}  \\
        {\bb_2}_j & = -\bc^\top_{j,\bbeta} \be_{\text{op}} = -\sum_{k=1}^{\numEvents} \left.\frac{\partial (\be)_k}{\partial {\bP_{j,\bbeta}}}\right|_{\text{op}} (\be_\text{op})_{k}.
    \end{aligned}
    \label{eq:RHSVecCumulative}
\end{equation}
where $\frac{\partial (\be)_k}{\partial {\bP_{\balpha}}_i}$ and $\frac{\partial (\be)_k}{\partial {\bP_{\bbeta}}_j}$ are the derivatives of the error term $(\be)_k$ with respect to the $i$/$j$-th component of the pose/map states.

Equations \eqref{eq:LHSMatCumulative} and \eqref{eq:RHSVecCumulative} allow us to accumulate the contribution of each event to the normal equations \eqref{eq:NormalEqs},
so that we can omit forming $\mJ_{\text{op}}$. 
The size of $\mA$ only depends on the dimension of state parameters, i.e., $(3 \numPoses + 2 \numPixels)^2$, which is significantly smaller than that of $\mJ_{\text{op}}$, i.e., $\numEvents \times (3 \numPoses + 2 \numPixels)$.

%\newpage
\subsection{Sensitivity and Ablation Analyses}
\label{sec:suppl:sensitivity}
We characterize the sensitivity of EMBA with respect to some of its parameters 
and also show the effect of a robust loss function. 
In the following, the map size is $1024 \times 512$ px, the initial rotations come from \cmaxw{}, 
and the sequence used is \bicycle{}. 

\subsubsection{Contrast Threshold}
Firstly, we run EMBA with varying values of $C=\{0.05,0.1,0.2,0.5,1.0\}$ in the loss function,
where $C=0.2$ is the true value for \bicycle{}.
We set $f=20$~Hz and $\eta = 5.0$.
Note that the value of $C$ affects the value of the PhE. 
Therefore, for a meaningful comparison, we use the PhE at $C=0.2$ as reference and calculate the equivalent PhE for the other $C$ values.
The results are presented in \cref{tab:sensitivity:contrast}.
The closer $C$ is to 0.2, the smaller the PhE.
The trials of $C=\{0.1,0.2\}$ achieve smaller rotation errors than the others.
Nevertheless, the trials of $C=\{0.05,0.5,1.0\}$ still show a strong refinement effect, in terms of both ARE and PhE 
(with respect to 1.69$^\circ$ ARE and $5.5 \cdot 10^5$ PhE, in \cref{tab:synth_traj_error,tab:synth_photo_error}),
which implies that EMBA is robust to the choice of $C$.
This is important in applications because the contrast thresholds of real event cameras are difficult to obtain 
and may vary greatly within the same dataset \cite{Stoffregen20eccv}.

\begin{table}[ht]
\centering
\caption{\label{tab:sensitivity:contrast} 
Sensitivity analysis on the camera's contrast threshold $C$.
Top: absolute rotation error (ARE), in RMSE form.
Bottom: equivalent squared photometric error.}
\adjustbox{max width=\linewidth}{
\setlength{\tabcolsep}{8pt}
\begin{tabular}{lrrrrr}
\toprule 
$C$  & 0.05 & 0.1 & 0.2 & 0.5 & 1.0 \\
\midrule
ARE [$^\circ$] & 1.193 & 0.899 & 0.923 & 0.966 & 1.341 \\
Equivalent PhE  [$\cdot 10^5$] & 3.024 & 2.956 & 2.956 & 2.968 & 3.030 \\
\bottomrule
\end{tabular}
}
\end{table}

\subsubsection{Weight of $\Ltwo$ Regularization}
We run EMBA with different values of $\eta=\{0,0.1,0.5,1.0,5.0,10.0,20.0\}$ while setting $C=0.2$ and $f=20$~Hz.
The results are shown in \cref{tab:sensitivity:weight}.
When $\eta = 0$, i.e., disabling the $\Ltwo$ regularization, the resulted gradient map is shown in \cref{fig:ablation_reg:disable}, where a few pixels dominate the optimization, thus suppressing the update of other pixels.
Meanwhile, it reports the worst ARE and PhE values among all $\eta$ values (\cref{tab:sensitivity:weight}).
This reveals that the $\Ltwo$ regularization is essential, and it effectively encourages a good convergence (like in \cref{fig:ablation_reg:enable}).
As $\eta$ increases from 0.1 to 5.0, both ARE and PhE decrease smoothly until they achieve their best values at $\eta = 5.0$; afterwards they increase with $\eta$.
Empirically, $\eta = 5.0$ is a good choice in most cases.

\begin{table}[ht]
\centering
\caption{\label{tab:sensitivity:weight} 
Sensitivity analysis on the weight of $\Ltwo$ regularization $\eta$.
}
\adjustbox{max width=\linewidth}{
\setlength{\tabcolsep}{8pt}
\begin{tabular}{lrrrrrrr}
\toprule 
$\eta$ & 0 & 0.1 & 0.5 & 1.0 & 5.0 & 10.0 & 20.0 \\
\midrule
ARE [$^\circ$] & 1.527 & 1.295 & 1.301 & 1.222 & 0.923 & 1.032 & 1.086 \\
Equivalent PhE [$\cdot 10^5$] & 3.160 & 3.053 & 3.049 & 3.032 & 2.956 & 3.015 & 3.071 \\
\bottomrule
\end{tabular}
}
\end{table}
\def\figWidth{0.49\linewidth}
\begin{figure}[ht]
  \centering
  \begin{subfigure}[c]{\figWidth}
         \centering
         \gframe{\includegraphics[width=\linewidth]{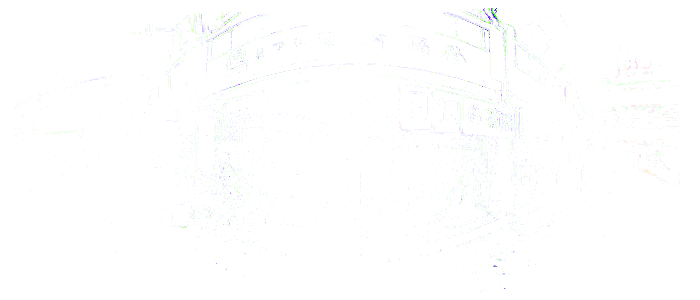}}
         \caption{$\eta = 0$.}
         \label{fig:ablation_reg:disable}
     \end{subfigure}
     \begin{subfigure}[c]{\figWidth}
         \centering
         \gframe{\includegraphics[width=\linewidth]{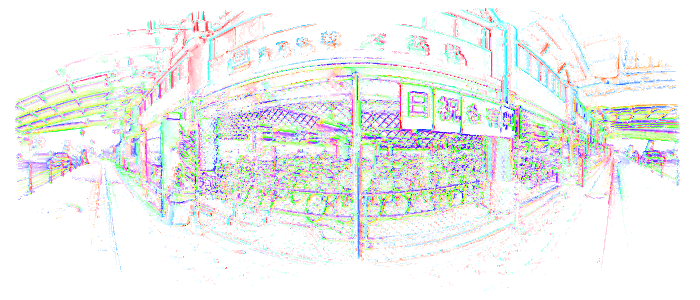}}
         \caption{$\eta = 5$.}
         \label{fig:ablation_reg:enable}
     \end{subfigure}
   \caption{Effect of $L^2$ regularization on the refined gradient map.}
   \label{fig:ablation_reg}
\end{figure}

\subsubsection{Robust Loss Function}
The formula of the Huber loss function is:
\begin{equation}
    \rho(u) = {\begin{cases}{u^{2}}&{\text{for }}|u| < \delta ,\\ 
\left(2|u|-\delta \right)\delta, &{\text{otherwise.}}\end{cases}}
\end{equation}

We apply it to each error term, $u = (\be)_k$, thus replacing the data-fidelity cost $\sum_k ((\be)_k)^2$ in \eqref{eq:NLLSGenericProblem}, \eqref{eq:NLLSRegProblem} by $\sum_k \rho((\be)_k)$.
In the experiments, we set $C=0.2$, $f=20$~Hz, $\eta = 5.0$ and $\delta=0.1$.

\Cref{tab:IRLS:synth,tab:IRLS:real} compare the Quadratic and Huber cost functions in terms of rotation error and PhE on synthetic and real-world data, respectively.
For a fair comparison, we present the squared PhE for both Quadratic and Huber loss.

\emph{ARE}:
On synthetic data, the Huber loss function results in slightly better rotation error than the Quadratic one in most trials, with only three exceptions.
All error differences are less than 0.35 degrees.
On real-world data it is hard to analyze the impact of the Huber loss function on rotation accuracy
due to the inherent evaluation problems (explained at the beginning of \cref{sec:experim:real}).

\emph{PhE}:
On the other hand, the refined PhE of the Huber loss is a little bigger than that of the Quadratic loss on most synthetic and real-world sequences.
This is a predictable result, because the objective function of the Huber loss has changed to a new ``reweighted'' squared PhE,
where the weights of the outliers are reduced.

\begin{table*}
\centering
\caption{\label{tab:IRLS:synth} 
Absolute rotation RMSE [deg] (ARE)
and squared photometric error [$\times 10^6$] (PhE)
on \emph{synthetic} sequences \cite{Guo24tro} (Schur solver, 1024 $\times$ 512 px map).}
\adjustbox{max width=\linewidth}{
\setlength{\tabcolsep}{3pt}
\begin{tabular}{ll*{11}{S[table-format=1.3,table-number-alignment=center]}}
\toprule
&& \multicolumn{3}{c}{\esmt{}} &\emptycol& \multicolumn{3}{c}{\cmaxgae{}} &\emptycol& \multicolumn{3}{c}{\cmaxw{}}\\
\cmidrule(l{2mm}r{2mm}){3-5}
\cmidrule(l{2mm}r{2mm}){7-9}
\cmidrule(l{2mm}r{2mm}){11-13}
&Sequence & \text{before} & \text{Quad} & \text{Huber} 
&\emptycol& \text{before} & \text{Quad} & \text{Huber}
&\emptycol& \text{before} & \text{Quad} & \text{Huber} \\
\midrule
%%%
\multirow{6}{*}{\begin{turn}{90}
ARE
\end{turn}}
&playroom & 5.861 & 6.094 & 6.147 & \emptycol & 4.628 & 4.419 & 4.316 & \emptycol & 3.223 & 2.856 & 2.786 \\
&bicycle & 1.466 & 1.182 & 1.012 & \emptycol & 1.651 & 1.496 & 1.413 & \emptycol & 1.69 & 0.923 & 0.973 \\
&city & 1.692 & 1.675 & 1.393 & \emptycol & \novalue & \NA{} & \NA{} & \emptycol & 1.532 & 0.973 & 0.94 \\
&street & 3.441 & 3.456 & 3.234 & \emptycol & \novalue & \NA{} & \NA{} & \emptycol & 0.965 & 0.744 & 0.744 \\
&town & 4.322 & 4.4 & 4.233 & \emptycol & 4.656 & 4.534 & 4.44 & \emptycol & 1.905 & 0.858 & 1.212 \\
&bay & 2.5 & 2.412 & 2.295 & \emptycol & \novalue & \NA{} & \NA{} & \emptycol & 1.797 & 1.409 & 1.391 \\
%%%
\midrule
%%%
\multirow{6}{*}{\begin{turn}{90}
PhE
\end{turn}}
&playroom & 0.347 & 0.226 & 0.26 & \emptycol & 0.35 & 0.186 & 0.212 & \emptycol & 0.326 & 0.151 & 0.183\\
&bicycle & 0.515 & 0.299 & 0.319 & \emptycol & 0.526 & 0.309 & 0.342 & \emptycol & 0.552 & 0.295 & 0.325 \\
&city & 2.623 & 2.126 & 2.185 & \emptycol & \novalue & \NA{} & \NA{} & \emptycol & 2.714 & 1.978 & 2.113 \\
&street & 1.821 & 1.518 & 1.504 & \emptycol & \novalue & \NA{} & \NA{} & \emptycol & 1.895 & 1.336 & 1.433 \\
&town & 1.88 & 1.514 & 1.618 & \emptycol & 1.904 & 1.537 & 1.646 & \emptycol & 1.917 & 1.425 & 1.552 \\
&bay & 2.259 & 1.961 & 1.954 & \emptycol & \novalue & \NA{} & \NA{} & \emptycol & 2.303 & 1.827 & 1.976 \\
%%%
\bottomrule
\end{tabular}
}
\end{table*}
\begin{table*}
\centering
\caption{\label{tab:IRLS:real} 
Absolute rotation RMSE [deg] (ARE)
and squared photometric error [$\times 10^6$] (PhE)
on \emph{real} sequences \cite{Mueggler17ijrr} 
(Schur solver, 1024 $\times$ 512 px map).}
\adjustbox{max width=\linewidth}{
\setlength{\tabcolsep}{3pt}
\begin{tabular}{ll*{11}{S[table-format=1.3,table-number-alignment=center]}}
\toprule
&& \multicolumn{3}{c}{\rtpt{}} &\emptycol& \multicolumn{3}{c}{\cmaxgae{}} &\emptycol& \multicolumn{3}{c}{\cmaxw{}}\\
\cmidrule(l{2mm}r{2mm}){3-5}
\cmidrule(l{2mm}r{2mm}){7-9}
\cmidrule(l{2mm}r{2mm}){11-13}
&Sequence & \text{before} & \text{Quad} & \text{Huber} 
&\emptycol& \text{before} & \text{Quad} & \text{Huber}
&\emptycol& \text{before} & \text{Quad} & \text{Huber} \\
\midrule
%%%
\multirow{4}{*}{\begin{turn}{90}
ARE
\end{turn}}
&shapes & 2.187 & 2.85 & 2.618 & \emptycol & 2.512 & 2.691 & 2.614 & \emptycol & 4.111 & 4.441 & 4.126 \\
&poster & 3.802 & 3.958 & 3.991 & \emptycol & 3.625 & 4.094 & 4.156 & \emptycol & 4.072 & 4.196 & 4.128 \\
&boxes & 1.743 & 2.319 & 2.256 & \emptycol & 2.018 & 2.4 & 2.317 & \emptycol & 3.224 & 2.866 & 2.921 \\
&dynamic & 2 & 2.285 & 2.4 & \emptycol & 1.698 & 2.004 & 1.967 & \emptycol & 3.126 & 2.791 & 2.8 \\
%%%
\midrule
%%%
\multirow{4}{*}{\begin{turn}{90}
PhE
\end{turn}}
&shapes & 0.675 & 0.374 & 0.516 & \emptycol & 0.609 & 0.383 & 0.503 & \emptycol & 0.575 & 0.361 & 0.497 \\
&poster & 4.692 & 2.583 & 2.881 & \emptycol & 5.03 & 3.069 & 3.298 & \emptycol & 4.368 & 2.579 & 2.87 \\
&boxes & 4.459 & 2.304 & 2.429 & \emptycol & 4.522 & 2.927 & 2.986 & \emptycol & 3.921 & 2.25 & 2.417 \\
&dynamic & 3.294 & 2.239 & 2.367 & \emptycol & 3.164 & 2.393 & 2.705 & \emptycol & 3.049 & 2.13 & 2.298 \\
%%%
\bottomrule
\end{tabular}
}
\end{table*}

In addition to \cref{tab:IRLS:synth,tab:IRLS:real}, we show a qualitative result here (more are available in the accompanying video).
\Cref{fig:robust} compares the refined maps produced by the quadratic and Huber loss functions.
The Huber panorama is similar and slightly sharper than the quadratic one.
\def\figWidth{0.49\linewidth}
\begin{figure}[ht]
     \centering
     \begin{subfigure}[c]{\figWidth}
         \centering
         \includegraphics[width=\linewidth]{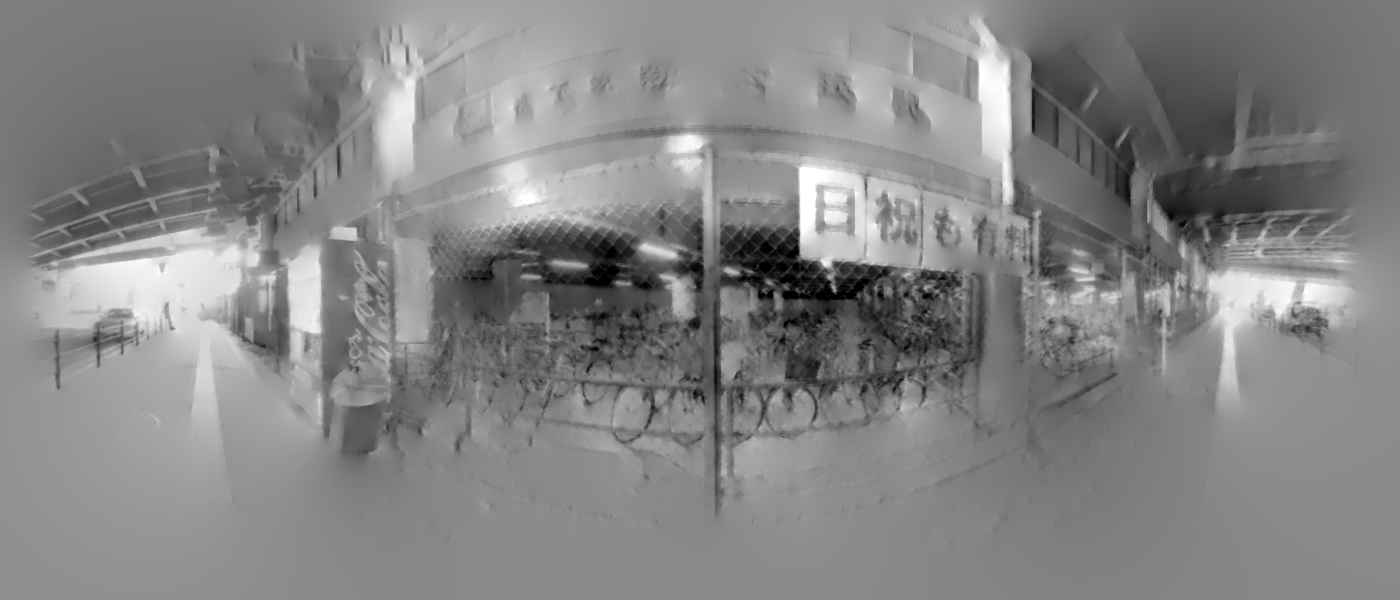}
         \caption{Quadratic.}
         \label{fig:robust:quad}
     \end{subfigure}\,%
     \begin{subfigure}[c]{\figWidth}
         \centering
         \includegraphics[width=\linewidth]{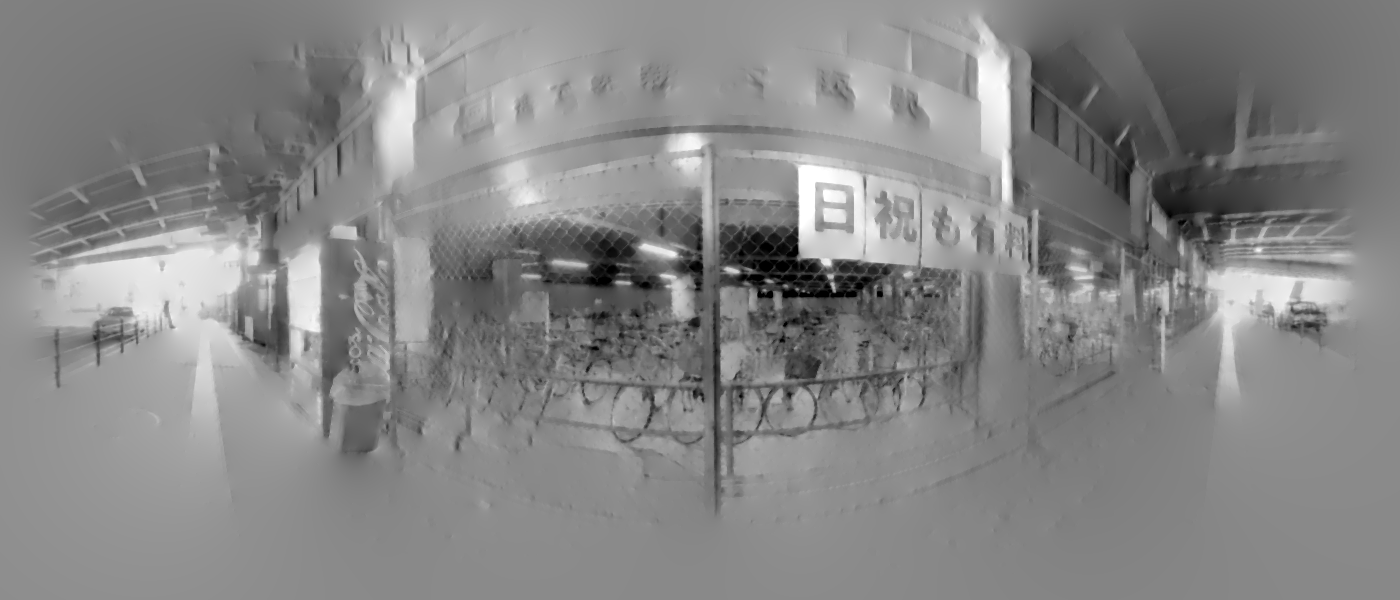}
         \caption{Huber.}
         \label{fig:robust:huber}
     \end{subfigure}\,%
        \caption{Effect of robust loss function. 
        Refined maps obtained with (a) Quadratic and (b) Huber loss functions. 
        (\bicycle{} sequence, initialized by \cmaxw{} trajectory).
        \label{fig:robust}}
\end{figure}

\subsubsection{Control Pose Frequency}
We run EMBA to refine the same initial rotations and maps, but varying the control pose frequency $f=\{10,20,50,100\}$~Hz. 
$C=0.2$ is set to its true value and $\eta = 5.0$.
The results are reported in \cref{tab:sensitivity:ctrl_pose_freq}.
It turns out that EMBA is also robust to the choice of $f$.
As $f$ grows from 10 to 50~Hz, both ARE and PhE decrease slightly and reach a minimum at $f=50$~Hz.
When $f$ is increased to $100$~Hz, the errors grow marginally, which implies that a too high $f$ does not lead to a better refinement.

\begin{table}[ht]
\centering
\caption{\label{tab:sensitivity:ctrl_pose_freq} 
Sensitivity analysis on the control pose frequency $f$.
}
\adjustbox{max width=0.75\columnwidth}{
\setlength{\tabcolsep}{8pt}
\begin{tabular}{lrrrr}
\toprule 
$f$~[Hz] & 10 & 20 & 50 & 100 \\
\midrule
ARE [$^\circ$] & 0.984 & 0.923 & 0.890 & 1.112 \\
PhE [$\cdot 10^5$] & 3.120 & 2.956 & 2.926 & 2.929 \\
\bottomrule
\end{tabular}
}
\end{table}

\subsection{Additional Discussion of the Experiments}

\subsubsection{Front-end failures}
In the experiments, four different front-end methods are used to initialize EMBA.
\rtpt{} fails on all synthetic sequences and \esmt{} fails on all real-world ones.
The explanation is as follows:
\rtpt{} loses track due to its limitation on the range of camera rotations that can be tracked.
It monitors the tracking quality during operation and stops updating the map when the quality decreases below a threshold, which offen happens if the camera’s FOV gets close to the left or right boundaries of the panoramic map.
The tracking failure of \esmt{} happens mostly when the camera changes the rotation direction abruptly.
We suspect it is due to the error propagation between the tracking and mapping threads.
Small errors in the poses or the map are amplified, corrupting the states and their uncertainty in the respective Bayesian filters.

\subsubsection{Camera translation in ECD datasets}
In \cref{sec:experim}, we mentioned that the four sequences from the ECD dataset \cite{Mueggler17ijrr} were recorded by a hand-held event camera, so the camera motion inevitably contains translations,
which affects all involved front-end methods as well as our BA approach.
\Cref{fig:suppl:cam_trans} displays the translational component of the GT poses provided by the mocap.
It shows that the magnitude of the translation grows, as time progresses and the speed of the motion increases.
We use the first part of the sequences, where the translational motion is still small (about less than 10 cm) for the desk-sized scenes.

\def\figWidth{0.48\linewidth}
\begin{figure*}[ht]
     \centering
    \begin{subfigure}[b]{\figWidth}
         \centering
         \includegraphics[width=\linewidth]{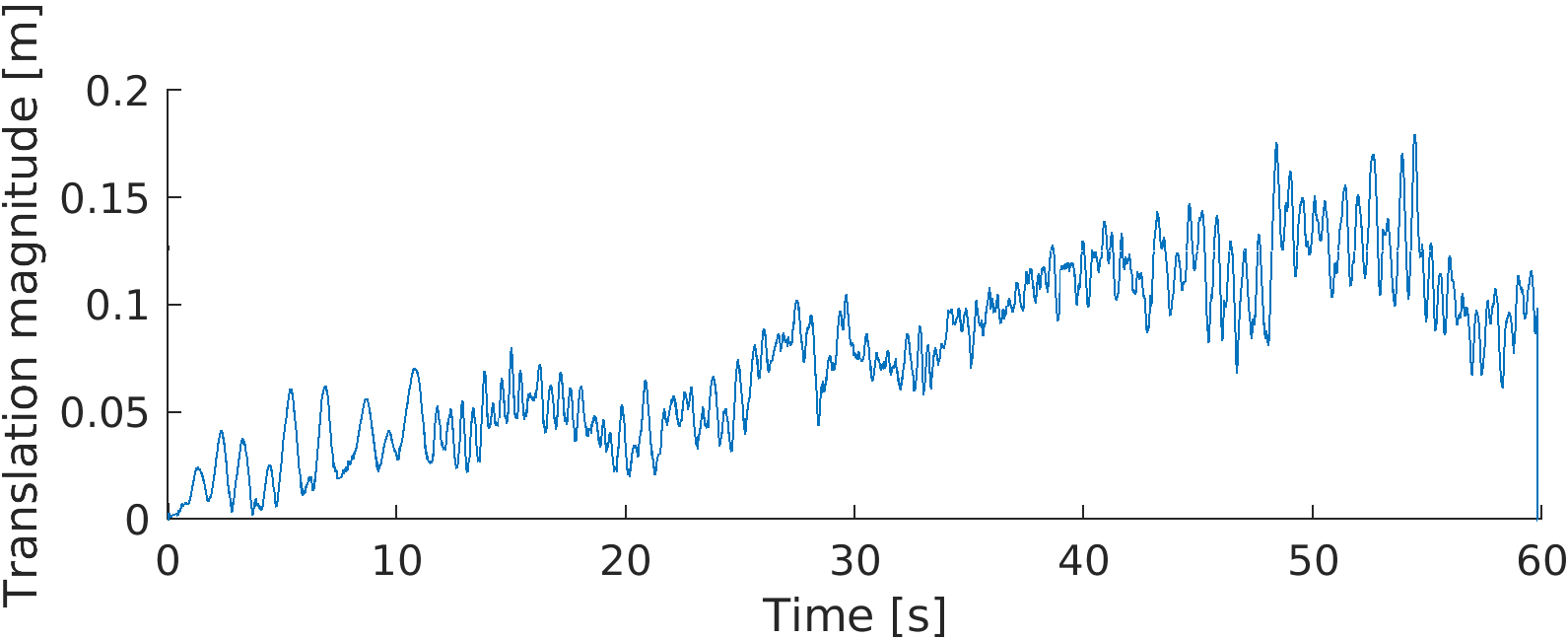}
         \caption{\shapes{}}
         \label{fig:suppl:cam_trans:shapes}
     \end{subfigure}\;\;\;\;
     \begin{subfigure}[b]{\figWidth}
         \centering
         \includegraphics[width=\linewidth]{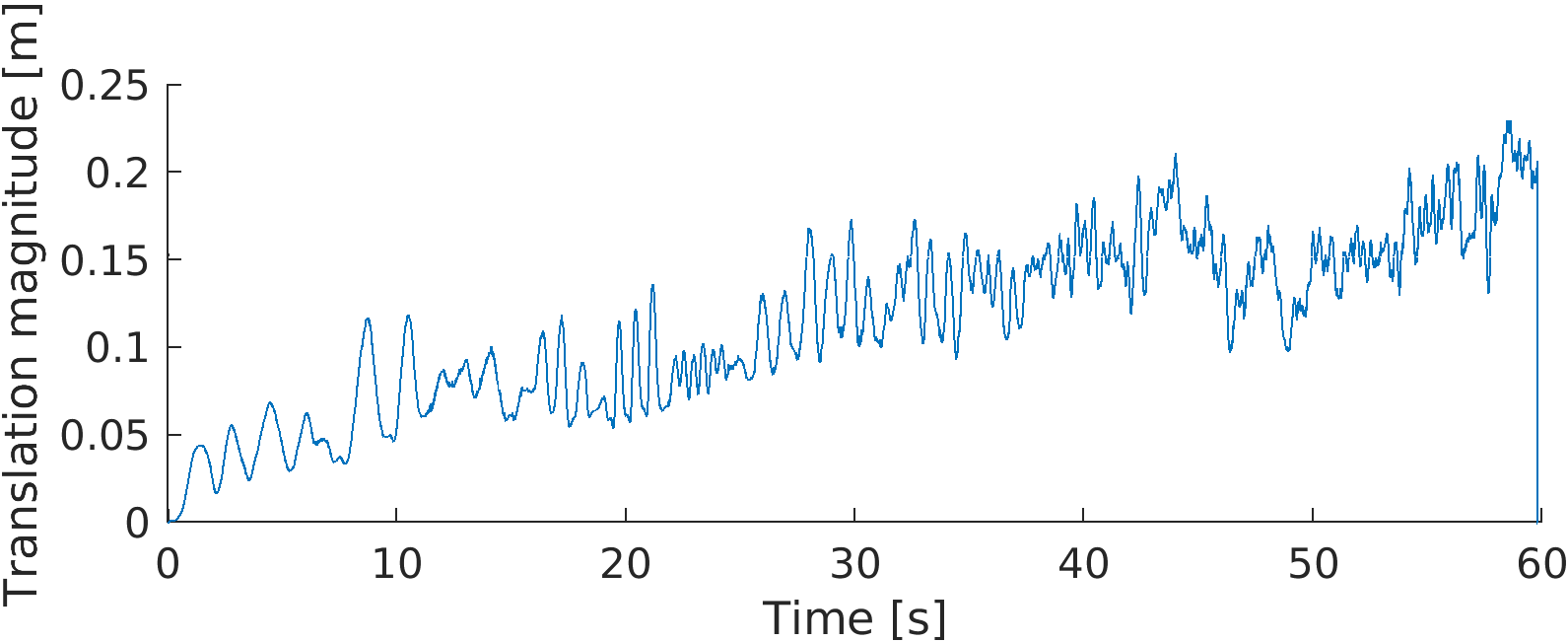}
         \caption{\poster{}}
         \label{fig:suppl:cam_trans:poster}
     \end{subfigure}
     \begin{subfigure}[b]{\figWidth}
         \centering
         \includegraphics[width=\linewidth]{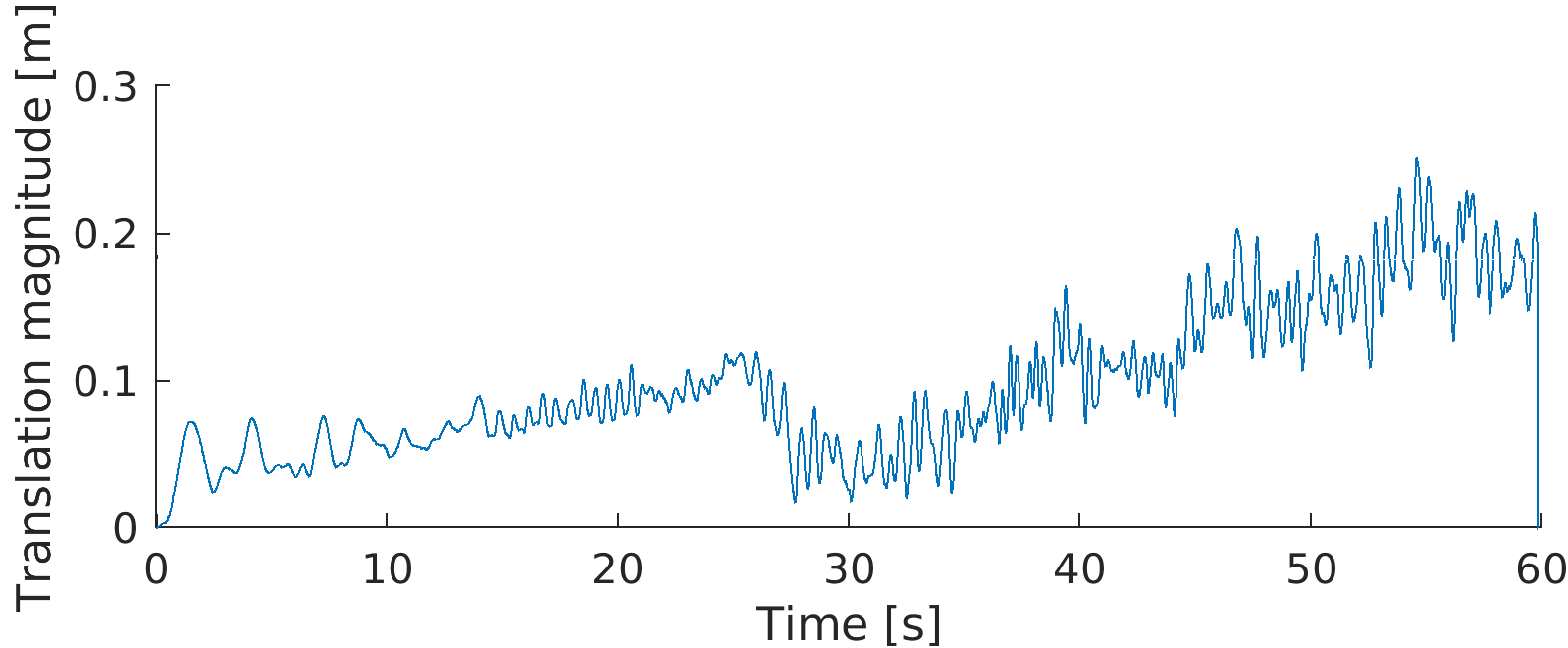}
         \caption{\boxes{}}
         \label{fig:suppl:cam_trans:boxes}
    \end{subfigure}\;\;\;\;
    \begin{subfigure}[b]{\figWidth}
         \centering
         \includegraphics[width=\linewidth]{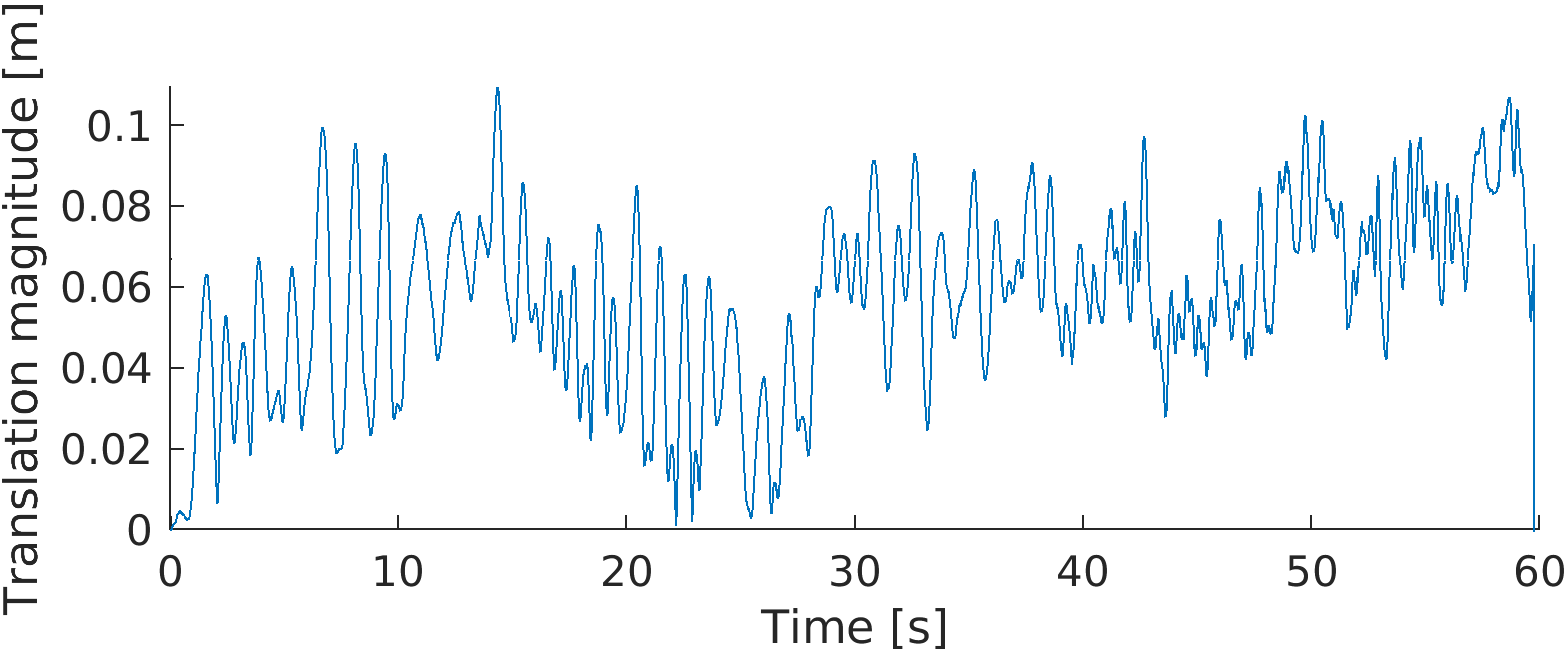}
         \caption{\dynamic{}}
         \label{fig:suppl:cam_trans:dynamic}
    \end{subfigure}
    \caption{From the motion capture system: groundtruth camera translation magnitude of the four ECD sequences \cite{Mueggler17ijrr}.
    }
    \label{fig:suppl:cam_trans}
\end{figure*}

\clearpage
\bibliographystyle{splncs04}
\bibliography{all,main}

\else 
% ---- Bibliography ----
\bibliographystyle{splncs04}
%\bibliography{all,main}

\clearpage
\setcounter{page}{1}

\fi 

\end{document}